\documentclass[sigconf]{acmart}
\makeatletter
\def\@ACM@checkaffil{
    \if@ACM@instpresent\else
    \ClassWarningNoLine{\@classname}{No institution present for an affiliation}%
    \fi
    \if@ACM@citypresent\else
    \ClassWarningNoLine{\@classname}{No city present for an affiliation}%
    \fi
    \if@ACM@countrypresent\else
        \ClassWarningNoLine{\@classname}{No country present for an affiliation}%
    \fi
}
\makeatother
\usepackage{array}
\usepackage{booktabs}
\usepackage{multirow}
\usepackage{subcaption}
\usepackage{makecell, booktabs, caption}
\usepackage{algpseudocode}
\usepackage[ruled,linesnumbered]{algorithm2e}
\usepackage{enumitem}
\usepackage{makecell}
\usepackage{diagbox}
\usepackage{graphicx}
\usepackage{bm}
\usepackage{makecell}
\usepackage{color}
\usepackage{balance}

\AtBeginDocument{%
  \providecommand\BibTeX{{%
    \normalfont B\kern-0.5em{\scshape i\kern-0.25em b}\kern-0.8em\TeX}}}

\copyrightyear{2023}
\acmYear{2023}
\setcopyright{acmlicensed}
\acmConference[CIKM '23] {Proceedings of the 32nd ACM International Conference on Information and Knowledge Management}{October 21--25, 2023}{Birmingham, United Kingdom.}
\acmBooktitle{Proceedings of the 32nd ACM International Conference on Information and Knowledge Management (CIKM '23), October 21--25, 2023, Birmingham, United Kingdom}
\acmPrice{15.00}
\acmISBN{979-8-4007-0124-5/23/10}
\acmDOI{10.1145/3583780.3614837}
\begin{document}
\title{Deep Task-specific Bottom Representation Network for Multi-Task Recommendation}

\author{Qi Liu}
\authornote{This work was done when the author Qi Liu was at Alibaba Group for intern.}
\affiliation{%
  \institution{University of Science and Technology of China}
}
\email{qiliu67@mail.ustc.edu.cn}

\author{Zhilong Zhou}
\affiliation{%
  \institution{Alibaba Group}
}
\email{zhilong.zhou1996@gmail.com}

\author{Gangwei Jiang}
\affiliation{%
  \institution{University of Science and Technology of China}
}
\email{gwjiang@mail.ustc.edu.cn}

\author{Tiezheng Ge}
\authornote{Corresponding author: Tiezheng Ge}
\affiliation{%
  \institution{Alibaba Group}
}
\email{tiezheng.gtz@alibaba-inc.com}

\author{Defu Lian}
\affiliation{%
  \institution{University of Science and Technology of China}
}
\email{liandefu@ustc.edu.cn}

\renewcommand{\shortauthors}{Qi Liu, Zhilong Zhou, Gangwei Jiang, Tiezheng Ge*, \& Defu Lian}

\begin{abstract}
Neural-based multi-task learning (MTL) has gained significant improvement, and it has been successfully applied to recommendation system (RS). Recent deep MTL methods for RS (e.g. MMoE, PLE) focus on designing soft gating-based parameter-sharing networks that implicitly learn a generalized representation for each task. However, MTL methods may suffer from performance degeneration when dealing with conflicting tasks, as negative transfer effects can occur on the task-shared bottom representation. This can result in a reduced capacity for MTL methods to capture task-specific characteristics, ultimately impeding their effectiveness and hindering the ability to generalize well on all tasks. In this paper, we focus on the bottom representation learning of MTL in RS and propose the Deep Task-specific Bottom Representation Network (DTRN) to alleviate the negative transfer problem. DTRN obtains task-specific bottom representation explicitly by making each task have its own representation learning network in the bottom representation modeling stage. Specifically, it extracts the user's interests from multiple types of behavior sequences for each task through the parameter-efficient hypernetwork. To further obtain the dedicated representation for each task, DTRN refines the representation of each feature by employing a SENet-like network for each task. The two proposed modules can achieve the purpose of getting task-specific bottom representation to relieve tasks' mutual interference. Moreover, the proposed DTRN is flexible to combine with existing MTL methods. Experiments on one public dataset and one industrial dataset demonstrate the effectiveness of the proposed DTRN. 
\end{abstract}





\maketitle

\section{Introduction}
Recommendation system (RS), which aims to provide preferred candidates (e.g., items, ads, videos, news, etc) upon user properties, has played a vital role in web applications. In the industrial scenario, RS needs to satisfy multiple objectives, such as click rate, conversion rate, and recommendation diversity. Multi-task learning (MTL) which trains a single network to produce multiple predictions for each objective meets the requirements. Thus MTL has attracted lots of attention in recommendation system~\cite{gu2020deep, tang2020progressive, ma2018modeling, xi2021modeling, liu2010unifying, pan2016mixed, wu2022multi, wen2020entire, zhao2019recommending, gao2019learning, lian2020lightrec, chen2021boosting, wu2023influence,lian2020personalized,chen2022fast} and make a success in lots of applications.

\begin{table}[h]
    \renewcommand\arraystretch{0.85}
    \caption{The average times of target item appearing in each behavior sequence for each task.}\label{tab:motivation_tim}
    \vspace{-0.3cm}
    \begin{tabular}{cl|c|c|c|c}
    \toprule
    \multicolumn{2}{c|}{Method}             & \multicolumn{1}{c}{watch}              & \multicolumn{1}{c}{click}             & \multicolumn{1}{c}{entering}             & \multicolumn{1}{c}{conversion} \\ 
    \midrule
    \multicolumn{2}{c|}{watch\_seq}         & \multicolumn{1}{c}{1.24}             & \multicolumn{1}{c}{1.01}            & \multicolumn{1}{c}{0.83}               & \multicolumn{1}{c}{0.69}  \\
    \multicolumn{2}{c|}{non\_watch\_seq}    & \multicolumn{1}{c}{0.60}             & \multicolumn{1}{c}{0.70}            & \multicolumn{1}{c}{0.61}               & \multicolumn{1}{c}{0.42}  \\ 
    \multicolumn{2}{c|}{click\_seq}         & \multicolumn{1}{c}{0.07}             & \multicolumn{1}{c}{0.13}            & \multicolumn{1}{c}{0.09}               & \multicolumn{1}{c}{0.07}  \\ 
    \multicolumn{2}{c|}{non\_click\_seq}    & \multicolumn{1}{c}{1.76}             & \multicolumn{1}{c}{1.49}            & \multicolumn{1}{c}{1.30}               & \multicolumn{1}{c}{1.06}  \\
    \multicolumn{2}{c|}{entering\_seq}      & \multicolumn{1}{c}{1.56}             & \multicolumn{1}{c}{2.00}            & \multicolumn{1}{c}{2.57}               & \multicolumn{1}{c}{1.30} \\ 
    \bottomrule
    
    \end{tabular}
\end{table}

\begin{figure*}
    \centering
    \includegraphics[width=0.82\textwidth]{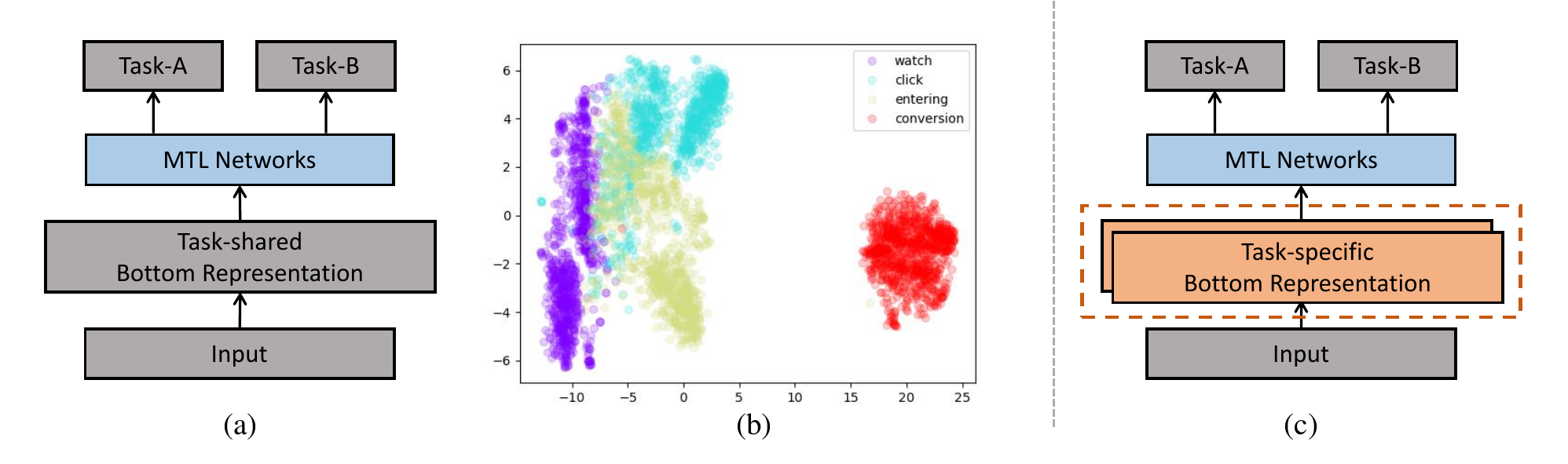}
    \vspace{-0.5cm}
    \caption{Existing MTL methods focus on the network structure (blue rectangles in (a)) with their weakness in obtaining task-specific representation (b). We focus on the neglected bottom representation learning (orange rectangles in (c)).}
    \label{fig:motivation_case}
\end{figure*}

The architecture of MTL methods in RS usually follows the paradigm as shown in Figure~\ref{fig:motivation_case}(a). These methods take the task-shared bottom representation as input and extract the task-specific representation through the gating-based parameter-sharing networks. Then, task-specific tower networks predict probabilities of actions (e.g. watch, click, and entering) based on task-specific representations. The benefits of the paradigm lie in two folds. First, it boosts multi-task performance by sharing information among related tasks. Second, it provides a highly compact and efficient mechanism of modeling for training and severing multi-task prediction. 

However, MTL in RS still faces the challenge of performance degradation caused by negative transfer between tasks with conflicting correlations. During the optimization process, the conflicting gradients produced by multiple tasks will lead to compromised bottom representation. The compromised bottom representation will inevitably hurt the generalization of the whole network and cause a negative transfer. For example, gradients from the conversion task will emphasize the income-related feature embedding while gradients of the clicking task should mainly focus on the hobby-related feature embedding. At the same time, both tasks give gradients to tyrannize the other related feature embedding which leads to compromised representation. Hence, how to solve the potential conflict between multiple tasks and simultaneously boost all tasks' performance is the core problem in MTL research. There are two major research lines, including the design of optimization strategies and network architectures. Prior studies~\cite{kendall2018multi, chen2018gradnorm, liu2019end, lin2019pareto, yu2020gradient, wang2020gradient} proposed to modify the gradient magnitude and directions adaptively to alleviate the conflict between multiple tasks' gradients. For network architecture, existing works~\cite{caruana1997multitask, tang2020progressive,ma2018modeling, xi2021modeling,wen2020entire,wen2021hierarchically,gao2019learning, ma2019snr,ding2021mssm,chen2021boosting,zou2022automatic} devoted much effort to the parameter-sharing mechanism, which aimed at solving the negative transfer problem through well-designed network structures being derived from either human's insight or Automated Machine Learning. However, these methods ignore the importance of the task-shared bottom representation. A generalized bottom representation is an essential factor in boosting MTL performance~\cite{wang2022can}. What's more, we observe that gating-based parameter-sharing methods have a limited ability in eliminating the compromise of representation. As shown in Figure~\ref{fig:motivation_case}(b), we visualize the task-specific representations which are the output of Progressive Layered Extraction (PLE)~\cite{tang2020progressive}, and there exist overlaps between these representations, which would incur conflicting gradients for different tasks when learning task-specific knowledge. We are motivated to fuse the task information into the bottom representations to obtain task-specific bottom representation as shown in Figure~\ref{fig:motivation_case}(c). 

\begin{figure*}[htb]
    \centering
    \includegraphics[width=0.95\linewidth]{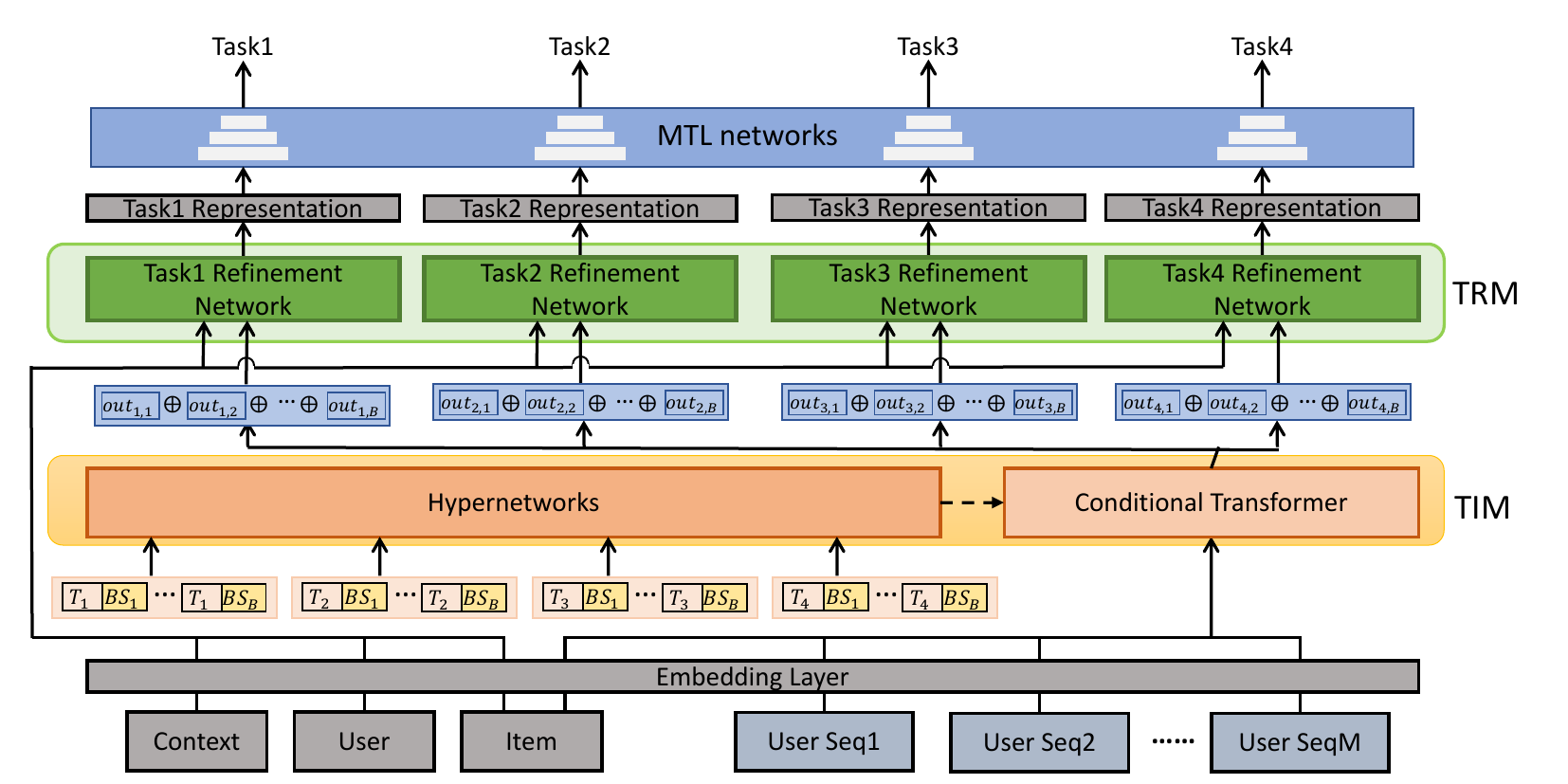}
    \caption{The overall framework of Deep Task-specific Representation Network(DTRN). DTRN consists of Task-specific Interest Module(TIM) and Task-specific Representation Refinement Module(TRM), which learns task-specific bottom representations.}
    \label{fig:model}
\end{figure*}

To address the above challenges, we propose a novel MTL model, called Deep Task-specific Bottom Representation Network (DTRN). As shown in Figure~\ref{fig:model}, it consists of the embedding layer, the task-specific interest module (TIM), the task-specific representation refinement module (TRM), and MTL Networks. Observing that various types of behavior sequences reflect different interests~\cite{grbovic2018real}, we are motivated to extract the task-specific interests with the multiple types of behaviors. As shown in Table~\ref{tab:motivation_tim}, we compute the average occurrence times of the target item's category in the user's each behavior sequence for each task. The result shows the correlation difference between different tasks and behavior sequences. This demonstrates the fine-grained influence difference of different types of historical behaviors on the user's current behavior. Instead of employing an independent behavior sequence modeling module for each task and behavior sequence pair, we innovatively design TIM based on the hypernetwork~\cite{ha2016hypernetworks}. In TIM, the number of behavior sequence modeling modules does not increase with the number of tasks and behavior sequences, so it is parameter-efficient and less prone to overfitting. Specifically, the hypernetwork takes the task and behavior embeddings as input to generate dynamic parameters, which will be injected into the base behavior sequence modeling module to produce the task-specific behavior sequence modeling module. Moreover, feature representation should be adaptive to different contexts and has various importance to different tasks. Additionally, to get the task-related context-aware representation namely task-specific bottom representation, we design a task-specific representation refinement module, consisting of multiple task-specific refinement networks to generate context-aware feature representation for each task. The task-specific bottom representation provides more choices for each task to synthesize gradients from other tasks and thus avoid the negative transfer. 

The main contributions of this paper are summarized as follows:
\begin{itemize}[leftmargin=*]
    \item We make a first attempt to achieve task-specific bottom representation learning in MTL models. We propose the Deep Task-specific Representation Network (DTRN) to replace the task-shared bottom representations in the MTL recommendation, which efficiently learns the task-specific bottom representation, and thus alleviates the negative transfer issues.
    \item We design the Task-specific Interest Module (TIM) to extract the user interest from multiple types of behavior sequences for each task, and the Task-specific Representation Refinement Module (TRM) to obtain the context-aware representation of each task, with the aim of achieving task-specific bottom representations for better generalization.
    \item We conduct offline experiments on both industrial and public datasets to verify the effectiveness of the proposed method DTRN, which achieves remarkable improvement w.r.t multiple business metrics in the online A/B test experiment.
\end{itemize}

\section{Related work}

\subsection{MTL in Recommendation System}
Earlier factorization-based models~\cite{liu2010unifying,pan2016mixed} learn task-specific user (or item) embeddings by sharing the other item (or user) embedding. Regularization methods are used to make the learned task-specific embeddings discriminative. Such methods have limited expressive ability and cannot fully unleash the power of MTL. Recently, deep learning-based multi-task models~\cite{wu2022multi,gu2020deep,xi2021modeling} have boosted recommendation performance significantly. Hard-sharing~\cite{caruana1997multitask} methods exploit the shared bottom network to achieve knowledge transfer, which will cause negative transfer for less relevant tasks. Expert-sharing~\cite{tang2020progressive,ma2018modeling} methods use multiple experts and task-specific gating networks to realize weighted parameter sharing. These methods try to strengthen the dependency and resolve conflicts between tasks. Defined-sharing~\cite{xi2021modeling,ma2018entire,wen2020entire,wen2021hierarchically,gao2019learning} methods realize knowledge transfer through the human-designed attention mechanism. Learning-base sharing methods~\cite{ma2019snr,ding2021mssm,chen2021boosting,zou2022automatic} utilize the power of AutoML or sparsity to learn the parameter sharing at different granularities automatically. All these methods focus on the design of parameter sharing and ignore the representation learning problem. We think converting the task-agnostic representation to a task-specific representation benefits the MTR.

\subsection{Behavior Sequence Modeling}
Behavior sequence modeling is an integral part of state-of-the-art recommendation models. DIN~\cite{zhou2018deep} discovers that only part items in the behavior sequence are related to the target item, and uses an attention MLP to activate those related items. DIEN~\cite{zhou2019deep} utilizes the dynamic change of user interest hidden in the user behavior sequence and proposes a two-layer attention-GRU network to fit this property. SIM~\cite{pi2020search} adopts the approximate nearest neighbor search technology to process the user's ultra-long behavior sequence. Recently, inspired by the success of Transformer~\cite{vaswani2017attention} in NLP and CV, ~\cite{gu2020deep,chen2019behavior,zhou2018atrank,xie2021deep} try to exploit Self-Attention to model the items' relationship before performing local activation. DMT~\cite{gu2020deep} proposes using separate Transformer encoder-decoder networks to model multiple  behavior sequences. However, most existing methods focus on single-type behavior modeling, especially click behavior. In real-world recommendation systems, we can obtain multiple types of behavior sequences based on multiple types of user feedback. These diversified behavior sequences can portray the user character accurately, which is hard to achieve by only exploiting click behavior. Going one step further, we find that the correlations between  tasks and behavior sequences diff. There are better solutions than simply extracting a shared interest representation from each behavior sequence for all tasks. The proposed DTRN can extract task-specific interest representations from each behavior sequence in the multi-task and multi-behavior scenarios. 

\section{Method}

\subsection{Overview}
Suppose there are $T$ tasks, $N$ types of sparse ID features, and $M$ types of user behavior sequences. We represent the instance by $\{\mathbf{x}, \mathbf{y}\}$, where $\mathbf{x}=[x_1^s,...,x_N^ s, x_{N+1}^b,...,x_{N+M}^b]$, $\mathbf{y}=[y_1,y_2,...,y_T]$. Here $x_i^s$ denotes the sparse ID feature (e.g., user\_id, age, item\_id and etc), $x_i^b$ denotes the user behavior sequence (e.g., the chronological items clicked by user), and $y_i\in\{0,1\}$ is the label for each task. MTL in recommendation system aims at estimating the probability $P(y_i|\mathbf{x}),i\in\{1,...,T\}$ for each instance. We formulate the MTL in RS as the following Eq.~\eqref{mtl_eq}:
\begin{equation}
    \label{mtl_eq}
    P(y_i|\mathbf{x}) = F(\mathbf{x}), i\in\{1,...,T\},
\end{equation}
where $F$ is the MTL model.

DTRN takes $\mathbf{x}$ as input and transforms it into dense vector through the embedding layer. The TIM takes task type embedding, behavior sequence type embedding, and embeddings of behavior sequences as input to extract task-specific interest. The task-specific interest together with sparse ID embeddings contains the original information namely the context of each instance. Based on the context, TRM performs representation refinement by applying Multi-Layer Perception network to adjust the importance of each feature for each task respectively. The task-specific bottom representation generated by TRM can combine with  MTL networks (like PLE) to make predictions of multiple tasks.

\subsection{Embedding Layer}

\textbf{Sparse ID}. $x_i^s$ is the sparse ID feature like user\_id, age, item\_id, it will be first converted to one-hot vector $\mathbf{t_i^s} \in R^{K_i}$, where $K_i$ denotes the cardinality of feature $i$. $t_i^s[j] = 1$ if the value of $x_i^s$ is assigned index $j$ in all values. And then an embedding matrix $\mathbf{E_i^s} \in R^{K_i \times d}$ is used to get the embedding vector $\mathbf{e_i^s}=\mathbf{t_i^sE_i^s}$, where $d$ is the embedding dimension.

\textbf{Behavior Sequence}. In RS, users have rich and diverse behaviors, such as clicking items, sharing items, and purchasing items. Such behaviors can be represented by a variable-length sequence of items $\mathbf{x_i^b}=\langle S_{b_1}, S_{b_2},..., S_{b_{N_i}} \rangle$, where $N_i$ is the length of behavior sequence and $S_{b_k}$ is the identifier of interacted items (like item\_id). Firstly, $\mathbf{x_i^b}$ will be transformed to sparse matrix $\mathbf{t_i^b} \in R^{N_i \times K_i}$, where $K_i$ denotes the cardinalty of feature $i$. $t_i^b[k,j]=1$ if the value of $S_{b_k}$ is assigned index $j$ in all values. Embedding matrix $\mathbf{E_i^b} \in R^{K_i \times d}$ is used to get the behavior sequence embedding $\mathbf{e_i^b}=\mathbf{t_i^bE_i^b}$, where $d$ is the embedding dimension.

\textbf{Task and Behavior Sequence Type} There are multiple tasks and multiple behavior sequences in our setting. We assign each task and each type of behavior sequence an embedding vector as their identification. There is a task type embedding matrix $\mathbf{E_T} \in R^{T \times d}$ and the embedding of task-$i$ is $\mathbf{T_i} = E_T[i]$. The behavior sequence type embedding matrix is $\mathbf{E_B} \in R^{M \times d}$ and behavior sequence-$b$'s type embedding is $\mathbf{BS_b} = E_B[b]$, where $M$ is the number of behavior sequence types. Task type and behavior sequence type embedding will also sever as the input of DTRN.

\subsection{TIM: Task-Specific Interest Module}
As shown in Figure~\ref{fig:tim}, the proposed TIM contains two major sub-modules, Hypernetwork, and Conditional Transformer. (1) Hypernetwork uses task type embedding and behavior sequence type embedding as inputs to dynamically generate conditional parameters. The generated parameters are supposed to capture the relatedness between corresponding task and behavior sequence. (2) Conditional Transformer contains one base Transformer network which is applied for modeling of behavior sequence and is shared by all task and behavior sequence pairs. The conditional parameters generated by hypernetwork will be injected into the layer normalization layer of the base Transformer network to achieve the goal of task-specific interest extraction. TIM has the advantage of being parameter-efficient as the parameters only come from the base Transformer network and hypernetwork.

\subsubsection{Hypernetwork} 
The key principle of TIM is to control the unified behavior sequence modeling network producing specific interest for each task and behavior pair. To achieve this, we apply hypernetwork to acquire the task and behavior sequence type embeddings and generate conditional parameters for the task and behavior sequence pair. These parameters will act as additional scale and shift parameters to the layer normalization layer in Transformer to generate the fine-grained user's interest hidden in the behavior sequence towards the specific task. Specifically, we adapt two hypernetworks to generate conditional scale and shift parameters respectively. We implement hypernetwork with a two-layer Multi-Layer Perceptron (MLP) and use ReLU as the hidden layer's activation function. 
The process is as the following Eq.~\eqref{hyper_param}: 
\begin{equation}
    \label{hyper_param}
    \gamma^l_{t,b}=MLP_{\theta^l_{\gamma}}(\mathbf{T_i, BS_b}), \beta^l_{t,b}=MLP_{\theta^l_{\beta}}(\mathbf{T_i, BS_b}),
\end{equation}
where $l$ represents the position of layer normalization in Transformer. $t$ and $b$ are the type indicator of task and behavior sequence. $\theta$ is the parameter of MLP. $\gamma^l$ and $\beta^l$ without subscript are the original scale and shift parameters of the layer normalization.

\subsubsection{Conditional Transformer}
With the success of Transformer in various scenarios, we take it as our base network to extract interest from behavior sequence. However, previous Transformer models can only handle one single task at a time, which is not suitable for the MTL. To this end, we propose a new architecture called Conditional Transformer, which learns the correlation and difference among various behavior sequences for different tasks in one unified model with conditional layer normalization. The components of Conditional Transformer are as follows:
 
\textbf{Multi-Head Self-Attention (MHSA):} To capture the relationships among item pairs from different subspaces, we apply the multi-head self-attention. MHSA first projects the sequence embedding into the query $Q$, key $K$, and value $V$ for $h$ times with different projection matrices. Then, attention mechanism uses the $Q$ and $V$ to compute the relatedness scores between all item pairs in the sequence. The calculated relatedness scores are applied to make a weighted sum of the values $V$ to get new sequence representation. Finally, the output of multiple heads will be concatenated together to get the final result through the projection matrix $W^O$. The MHSA can be expressed as follows: 
\begin{equation}
    MHSA(X)=concat(head_1,...,head_h)W^O,
\end{equation}
\begin{equation}
    head_i=Softmax(\frac{XW^Q_i({XW^K_i})^T}{\sqrt{d'}})XW^V_i,
\end{equation}
where h is the number of heads, $W^Q_i, W^K_i, W^V_i \in R^{d \times d'}$, $W^O \in R^{d \times d}$. And $d$ and $d'$ is the dimension of the input and weight vectors while $d'=\frac{d}{h}$.

\textbf{Point-wise Feed-Forward Networks (FFN):} To increase the representation ability of Transformer, FFN will be applied to perform nonlinear transform after the MHSA. FFN is a two-layer MLP with the ReLU activation in between, which is as Eq.~\ref{ffn}.
\begin{equation}
\label{ffn}
    FFN(X)=ReLU(XW_1+b_1)W_2+b_2,
\end{equation}
where $W_1 \in  R^{d \times d_f}$, $b_1 \in R^{d_f}$, $b_2 \in R^{d}$,  $W_2 \in R^{d_f \times d}$ and $d_f$ is the dimension of hidden state.

\textbf{Conditional Layer Normalization (CLN):} Inspired by~\cite{mahabadi2021parameter,tay2021hypergrid}, to allow base Transformer to adapt to each task-behavior sequence pair, the generated conditional scale parameter $\gamma^l_{t,b}$ and conditional shift parameter $\beta^l_{t,b}$ derived from the hypernetwork based on the task and behavior sequence type embeddings will be integrated to the layer normalization, which can be described as follows:
\begin{equation}
    \label{cln}
    CLN^l_{t,b}(X)=\gamma^l_{t,b} \cdot \gamma^l \cdot \frac{X-\mu}{\delta} + \beta^l_{t,b} + \beta^l
\end{equation}
where $l$ is the position of layer normalization, $t$ is the task type, $b$ is the behavior sequence type, $\mu$, $\delta$ is the mean and standard deviation of the input. The CLN empowers Transformer to have the ability to manipulate hidden representations by scaling them up or down, shifting them left or right, shutting them off, etc based on different conditional information with minimal parameters cost.

\begin{figure}[th]
    \centering
    \includegraphics[width=1.\columnwidth]{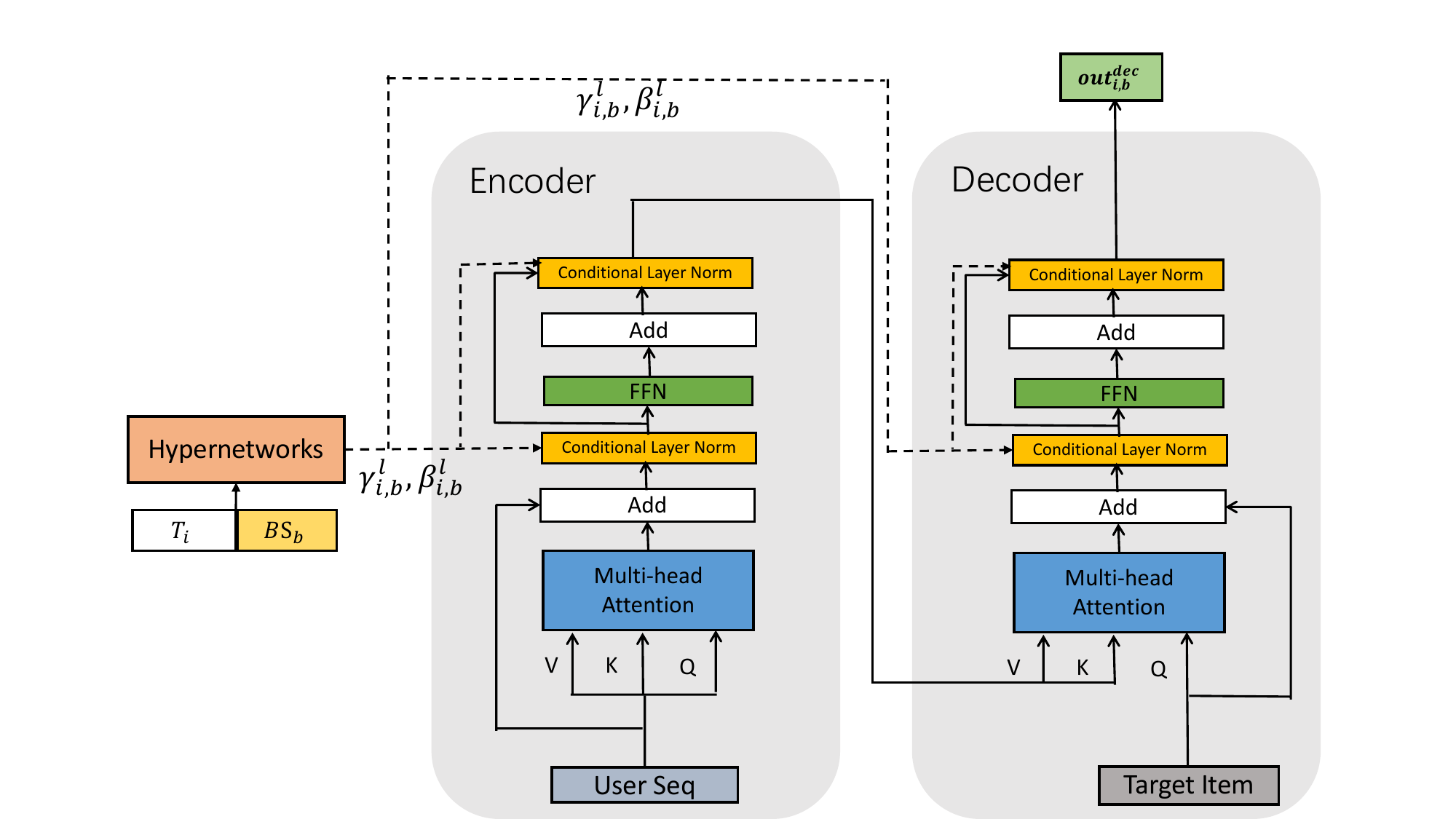}
    \caption{The network structure of TIM.}
    \label{fig:tim}
\end{figure}

\textbf{Components Integration:} As shown in Figure~\ref{fig:tim}, Conditional Transformer follows the encoder-decoder structure. The encoder has the ability to capture the relatedness of items by swapping information between items from different latent spaces with the help of MHSA, which can strengthen the representation of each item. The decoder also exploits the power of MHA to finish the local activation to get interest by using the target item as query and the output of encoder as keys and values. MHA (Multi-head attention) is similar to MHSA, except that query comes from target item. Specifically, the encoder and decoder layer are both made up of the MH(S)A, the FFN, and the CLN. They both exploit residual connections around each sublayer. The integrated components can be expressed as Eq.~\eqref{conditional_transformer}:
\begin{equation}
 \label{conditional_transformer}
    \begin{split}
        &out^0_{t,b}=CLN_{t,b}(X+MHSA(X)),\\
        &out^{enc}_{t,b}=CLN_{t,b}(out^0_{t,b}+FFN(out^0_{t,b}),\\
        &out^3_{t,b}=CLN_{t,b}(e_{item}+MHA(e_{item},out^{enc}_{t,b})),\\
        &out^{dec}_{t,b}=CLN_{t,b}(out^3_{t,b}+FFN(out^3_{t,b})),
    \end{split}
\end{equation}
where $X=e_i^s, i \in [N+1,N+M]$ is the embedding of behavior sequence-i, $e_{item}$ is the embedding of target item. The final output $out^{dec}_{t,b}$ is the interest extracted from behavior sequence-i for task-t. As we have $T$ tasks and $M$ types of behavior sequences, we will get $T \times M$ such interests.

\textbf{Task-Specific Interest:} After getting $T \times M$ interests for each task and behavior sequence pair, we will concatenate interests grouped by task to obtain the task-specific interest as Eq.~\eqref{ts_interest}:
\begin{equation}
    \label{ts_interest}
    interest_i=concat(out^{dec}_{i,1},out^{dec}_{i,2},...,out^{dec}_{i,M}), i\in\{1,...,T\},
\end{equation}
where $interest_i$ is the task-$i$'s task-specific interest.

\subsection{TRM: Task-specific Representation Refinement Module}
The TRM takes sparse ID embeddings with task-specific interests as input and aims at refining the representation of each feature. The refinement has two advantages. First, it empowers feature representation to be adaptive under different contexts. For example, the feature \emph{food}'s representation should be related to fast food on Wednesday while should contain information about restaurants on Sunday. Second, it meets the requirements that different features have various importance for different tasks~\cite{huang2019fibinet}. We develop a SENet-like network to achieve the above advantages by assigning importance vector on the raw dense vector of all features for each task  based on the context. First, we concatenate sparse ID embeddings together with task-specific interest to get raw task-specific bottom representation $raw_i=[e^s_{1},e^s_{2},...,e^s_{N},interest_i]$.
The raw task-specific bottom representation contains the original information namely context about the instance and serves as input of each task's SENet-like network. The output of SENet-like network acts as refinement vector and will do element-wise multiply with raw task-specific bottom representation to perform task-related context-aware representation refinement to get the task-specific bottom representation. We implement SENet-like network through a two-layer MLP with ReLU as the hidden layer's activation and Sigmoid as the output layer's activation function. 
The whole process is as Eq.~\eqref{trm}:
\begin{equation}
    \label{trm}
    r_i = raw_i \odot refine_i,\, refine_i=MLP_i(raw_i),  
\end{equation}
where $r_i, i\in\{1,...,T\}$ is the task-specific bottom representation.

\subsection{Multi-Task Learning}
There are usually multiple objectives in an industrial recommendation system, such as click, like, and conversion. The ranking system should be able to learn and estimate multiple objectives and combines these estimations to compute a final preference ranking score. The proposed DTRN can combine with MTL methods in RS for multi-task learning.
With the help of TIM and TRM, we can get $T$ different task-specific bottom representation $r_i$. Individual $r_i$ will be fed into the MTL model parallelly. Then, the MTL model output the multi-task logits $[o_1,o_2,..,o_T]$, which is illustrated as Eq.~\eqref{mtl_learn}:
\begin{equation}
\label{mtl_learn}
    {[o_1,o_2,...,o_T]}=MTL\_model(r_1, r_2,...,r_T),
\end{equation}
The loss function for the task-$t$ is $L_t$. The total loss $L$ is the sum of losses in multi-task:
\begin{equation}
        L =\sum_{t=1}^TL_t(\hat{y}_t,y_t), \hat{y}_t=\sigma(o_t),
\end{equation}
where $y_t$ is the ground truth label, $\hat{y_t}$ is the prediction probability, and $\sigma$ denotes the sigmoid function. In our experiments, we set all $L_t$ binary cross entropy loss.

\section{Experiments Setup}
\subsection{Datasets}

\begin{table}[tbp]
 \centering
 \caption{The statistic of the E-commerce dataset.}
 \label{tab:ecomere_dataset}
\resizebox{0.93\linewidth}{!}{%
\begin{tabular}{c|c|c|c}
\toprule
\#Sample & \#Feature & \#User & \#Item\\
\midrule
376,197,451 & 116 & 59,677,310 & 41,343 \\
\midrule
\#Watch & \#Click & \#Entering & \#Conversion \\
\midrule
218,732,912 & 5,648,867 & 2,747,893 &  45,972 \\
\bottomrule
\end{tabular}
}
 \resizebox{0.93\linewidth}{!}{%
\begin{tabular}{c|c|c|c|c|c}
\toprule
      Sequence  & Watch & \begin{tabular}[c]{@{}c@{}}Non\\ Watch\end{tabular} & Click & \begin{tabular}[c]{@{}c@{}}Non\\ Click\end{tabular} & Entering \\ \midrule
avg length & 9.98     & 7.34            & 0.65       & 16.96           & 22.80          \\
max length & 50         & 50              & 50         & 50              & 100           \\ \bottomrule
\end{tabular}
 }
\end{table}

\begin{table}[tbp]
 \centering
 \caption{The statistic of the Tenrec dataset.}
 \label{tab:tenrec_dataset}
 \resizebox{0.87\linewidth}{!}{%
\begin{tabular}{c|c|c|c}
\toprule
\#Sample & \#Feature & \#User & \#Item \\
\midrule
117,551,246 & 16 & 993,554 & 821,744 \\
\midrule
\#Click & \#Follow & \#Like & \#Share \\
\midrule
28,358,705 & 176,496 & 2,227,004 & 243,310 \\
\bottomrule
\end{tabular}
 }
 \resizebox{0.87\linewidth}{!}{%
\begin{tabular}{c|c|c|c|c|c}
\toprule
      Sequence & Click & Follow & Like & Share & Negative \\ \midrule
avg length & 11.62     & 0.12            & 0.87       & 0.21           & 9.8          \\
max length & 50         & 10              & 20         & 10              & 50           \\ \bottomrule
\end{tabular}
 }
\end{table}

\textbf{E-commerce Dataset:} We collect traffic logs from the display advertising system in Alibaba's micro-video e-commerce platform. One-month samples are used for training and samples of the following day are for testing. The size of the training and testing set is about 361 million and 15 million respectively. We need to predict four tasks: \emph{watch, click, entering, and conversion}. Task \emph{watch} means whether the user will watch the recommended micro-video for more than three seconds. Task \emph{click} measures the probability that user touches the "like" button during watching. Task \emph{entering} indicates that user goes to the detailed page of the merchandise during watching. Task \emph{conversion} expresses that the user purchases the merchandise finally. There are five behavior sequences that are aggregated based on different types of user behavior. Detailed statistics about the E-commerce dataset are shown in Table~\ref{tab:ecomere_dataset}. 

\textbf{Tenrec Dataset:}\footnote{\url{https://github.com/yuangh-x/2022-NIPS-Tenrec}} Tenrec dataset is also the collected user logs from the industrial platform. We conduct experiments on a subset named \textbf{QK-Video-1M} which contains one million users' behavior logs. We also need to predict four tasks in this dataset: \emph{click, follow, like, and share}. \emph{click} means the user clicks the displayed video and begins watching. \emph{follow} indicates the user follows the video's uploader. \emph{like} means the user touches the "like" button, and \emph{share} expresses the user shares the video with his/her friends on social media. There are also five types of behavior sequences. The negative behavior sequence consists of videos that are skipped by the user. Statistics of the Tenrec dataset are shown in Table~\ref{tab:tenrec_dataset}.

\subsection{Baselines}
We choose baselines from three aspects. First, we choose methods focusing on behavior sequence modeling, including \textbf{YouTubeNet}~\cite{covington2016deep}, \textbf{DIN}~\cite{zhou2018deep}, \textbf{DIEN}~\cite{zhou2019deep}, \textbf{BST}~\cite{chen2019behavior}, \textbf{ATRank}~\cite{zhou2018atrank}, \textbf{DeepFeed}~\cite{xie2021deep}, \textbf{DMT}~\cite{gu2020deep} which extract user's interest vector from a single type of behavior sequence or multiple types of behavior sequences. Second, we compare with existing MTL methods in RS, including \textbf{MMoE}~\cite{ma2018modeling}, \textbf{PLE}~\cite{tang2020progressive}, \textbf{M2M}~\cite{zhang2022leaving}, \textbf{AITM}~\cite{xi2021modeling}. Third, \textbf{SamllHeads}~\cite{wang2022can} which concentrates on improving the generalization of the task-shared bottom representation.

\subsection{Evaluation Metric}
Two widely used metrics AUC, and LogLoss are chosen. The AUC (Area Under the ROC Curve) measures the ranking accuracy. A higher AUC indicates better performance. The LogLoss measures the accuracy of the estimated probability depending on the ground-truth label. Even a slight improvement is considered a significant boost for the industry recommendation task, as it leads to a significant increase in revenue.

\subsection{Implementation Details}
For the details of experiments, there is a slight difference between the two datasets. For the E-commerce Dataset, there are causation relationships among tasks, which urges us to select AITM~\cite{xi2021modeling} as the MTL model combined with DTRN. One line of causality is \emph{watch->click}. Another line of causality is the \emph{watch->entering->conversion}. Specifically, the causality of \emph{watch->entering->conversion} means that the user will first watch the micro-video for more than three seconds, then goes to the detailed page of the merchandise and purchase the merchandise finally. As the tasks in the Tenrec dataset have no apparent causal relationship, we apply the MMoE~\cite{ma2018modeling} as the MTL model combined with DTRN. For baselines of focusing on a single type of behavior sequence, we choose the watch behavior sequence for E-commerce related experiments and the click behavior sequence for the Tenrec dataset. Notice, we still do multi-task learning for a single type of behavior sequence baselines. For the other baselines, we use all five types of behavior sequences in both two datasets. For the batch size, we set it to $2,048$ for both datasets and use Adam as the optimizer. The learning rate is $1e-4$ and $1e-3$ for E-commerce and Tenrec respectively, and one-epoch training is applied due to the one-epoch phenomenon~\cite{zhang2022towards}. We run all experiments through XDL2\cite{zhang2022picasso}.

\begin{table*}[htbp]
    \renewcommand\arraystretch{0.85}
    \caption{Performance comparison of baselines on two datasets. The best result is in boldface and the second best is underlined. ** indicates that difference to the best baseline is statistically significant at 0.01 level, and * represents 0.05 level.}\label{tab:main_result}
    \begin{tabular}{c|c|cc|cc|cc|cc}
    \toprule 
        \multirow{2}{*}{Dataset}    & \multirow{2}{*}{Method} & \multicolumn{2}{c|}{Watch}                                      & \multicolumn{2}{c|}{Click}                                            & \multicolumn{2}{c|}{Entering}                                     & \multicolumn{2}{c}{Conversion}                                \\ 
                                                              & & AUC                                     & LogLoss               & \multicolumn{1}{c}{AUC}                   & LogLoss                   & \multicolumn{1}{c}{AUC}                   & LogLoss               & \multicolumn{1}{c}{AUC}                   & LogLoss           \\ \midrule 
        \multirow{11}{*}{E-commerce} & YouTubeNet             & \multicolumn{1}{c}{0.7238}              & 0.606855              & \multicolumn{1}{c}{0.7113}                & 0.077608                  & \multicolumn{1}{c}{0.8171}                & 0.043184              & \multicolumn{1}{c}{0.8647}                & 0.001475          \\ 
                                    & DIN                     & \multicolumn{1}{c}{0.7245}              & 0.606214              & \multicolumn{1}{c}{0.7087}                & 0.077748                  & \multicolumn{1}{c}{0.8171}                & 0.043144              & \multicolumn{1}{c}{0.8636}                & 0.001488          \\ 
                                    & DIEN                    & \multicolumn{1}{c}{0.7257}              & 0.605454              & \multicolumn{1}{c}{0.7110}                & 0.077597                  & \multicolumn{1}{c}{0.8173}                & 0.043134              & \multicolumn{1}{c}{0.8650}                & 0.001478          \\ 
                                    & ATRANK                  & \multicolumn{1}{c}{0.7316}              & 0.600047              & \multicolumn{1}{c}{0.7186}                & 0.077087                  & \multicolumn{1}{c}{0.8158}                & 0.043219              & \multicolumn{1}{c}{0.8484}                & 0.001539          \\ 
                                    & DFN                     & \multicolumn{1}{c}{0.7318}              & 0.600256              & \multicolumn{1}{c}{0.7230}                & 0.076919                  & \multicolumn{1}{c}{0.8232}                & 0.042808              & \multicolumn{1}{c}{\underline{0.8727}}    & 0.001466          \\ 
                                    & DMT (MMoE)              & \multicolumn{1}{c}{0.7338}              & 0.598775              & \multicolumn{1}{c}{\underline{0.7246}}    & 0.076762                  & \multicolumn{1}{c}{0.8237}                & 0.042764              & \multicolumn{1}{c}{0.8726}                & 0.001466          \\ 
                                    & PLE                     & \multicolumn{1}{c}{0.7340}              & 0.598761              & \multicolumn{1}{c}{0.7237}                & 0.076778                  & \multicolumn{1}{c}{0.8231}                & 0.042772              & \multicolumn{1}{c}{0.8693}                & 0.001472          \\
                                    & AITM                    & \multicolumn{1}{c}{0.7337}              & 0.598964              & \multicolumn{1}{c}{0.7245}                & 0.076755                  & \multicolumn{1}{c}{\underline{0.8240}}    & 0.042739              & \multicolumn{1}{c}{0.8691}                & 0.001475          \\
                                    & M2M                     & \multicolumn{1}{c}{\underline{0.7343}}  & \underline{0.598345}  & \multicolumn{1}{c}{0.7245}                & 0.076732                  & \multicolumn{1}{c}{0.8238}                & 0.042740              & \multicolumn{1}{c}{0.8674}                & 0.001478          \\
                                    & SmallHeads              & \multicolumn{1}{c}{0.7342}              & 0.598699              & \multicolumn{1}{c}{\underline{0.7246}}    & \underline{0.076709}      & \multicolumn{1}{c}{\underline{0.8240}}    & \underline{0.042726}  & \multicolumn{1}{c}{0.8673}                & \underline{0.001471}          \\ \cmidrule{2-10} 
                                    & DTRN                    & \multicolumn{1}{c}{\bm{$0.7346^{**}$}}                 & \bm{$0.59816^{**}$}      & \multicolumn{1}{c}{\bm{$0.7271^{**}$}}       & \bm{$0.076584^{**}$}                 & \multicolumn{1}{c}{\bm{$0.8283^{**}$}}       & \bm{$0.042496^{**}$}                         & \multicolumn{1}{c}{\bm{$0.8828^{**}$}}       & \bm{$0.001456^{**}$} \\ \bottomrule \toprule
         
        \multirow{2}{*}{Dataset}    & \multirow{2}{*}{Method} & \multicolumn{2}{c|}{Click}                                      & \multicolumn{2}{c|}{Follow}                                       & \multicolumn{2}{c|}{Like}                                         & \multicolumn{2}{c}{Share}                            \\ 
                                    &                         & AUC                                 & LogLoss               & \multicolumn{1}{c}{AUC}                   & LogLoss                   & \multicolumn{1}{c}{AUC}                   & LogLoss               & \multicolumn{1}{c}{AUC}                       & LogLoss           \\ \midrule 
        \multirow{11}{*}{Tenrec}    & YouTubeNet              & \multicolumn{1}{c}{0.9458}              & 0.189366              & \multicolumn{1}{c}{0.8860}                & 0.002852                  & \multicolumn{1}{c}{0.9289}                & 0.024425              & \multicolumn{1}{c}{0.9091}                    & 0.002369     \\
                                    & DIN                     & \multicolumn{1}{c}{0.9460}              & 0.189854              & \multicolumn{1}{c}{0.8863}                & 0.002897                  & \multicolumn{1}{c}{0.9271}                & 0.024079              & \multicolumn{1}{c}{0.9032}                    & 0.002375          \\ 
                                    & DIEN                    & \multicolumn{1}{c}{0.9466}              & 0.188511              & \multicolumn{1}{c}{0.8820}                & 0.002927                  & \multicolumn{1}{c}{0.9308}                & 0.023904              & \multicolumn{1}{c}{0.9074}                    & 0.002408          \\ 
                                    & ATRANK                  & \multicolumn{1}{c}{0.9467}              & 0.188962              & \multicolumn{1}{c}{0.8954}                & 0.002879                  & \multicolumn{1}{c}{0.9362}                & 0.023156              & \multicolumn{1}{c}{0.9198}                    & 0.002412          \\
                                    & DFN                     & \multicolumn{1}{c}{0.9504}              & 0.181401              & \multicolumn{1}{c}{0.9063}                & 0.002768                  & \multicolumn{1}{c}{0.9489}                & 0.021467              & \multicolumn{1}{c}{0.9244}                    & 0.002301          \\
                                    & DMT (MMoE)              & \multicolumn{1}{c}{0.9506}              & 0.180974              & \multicolumn{1}{c}{\underline{0.9152}}    & 0.002724                  & \multicolumn{1}{c}{0.9501}                & 0.021243              & \multicolumn{1}{c}{0.9279}                    & 0.002280          \\
                                    & PLE                     & \multicolumn{1}{c}{0.9508}              & 0.180515              & \multicolumn{1}{c}{0.9149}                & 0.002684                  & \multicolumn{1}{c}{0.9504}                & 0.021063              & \multicolumn{1}{c}{0.9283}                    & 0.002262          \\
                                    & AITM                    & \multicolumn{1}{c}{0.9507}              & 0.180837              & \multicolumn{1}{c}{0.9102}                & 0.002711                  & \multicolumn{1}{c}{0.9501}                & 0.021218              & \multicolumn{1}{c}{0.9252}                    & 0.002305          \\
                                    & M2M                     & \multicolumn{1}{c}{0.9507}              & 0.180715              & \multicolumn{1}{c}{0.9128}                & \underline{0.002645}      & \multicolumn{1}{c}{0.9506}                & \underline{0.021042}              & \multicolumn{1}{c}{\underline{0.9289}}        & \underline{0.002247}        \\
                                    & SmallHeads              & \multicolumn{1}{c}{\underline{0.9510}}  & \underline{0.180377}  & \multicolumn{1}{c}{0.9150}                & 0.002660                  & \multicolumn{1}{c}{\underline{0.9508}}    & 0.021304  & \multicolumn{1}{c}{0.9274}                    & 0.002252          \\ \cmidrule{2-10} 
                                    & DTRN                    & \multicolumn{1}{l}{\bm{$0.9516^*$}}  & \bm{$0.179113^*$}     & \multicolumn{1}{l}{\bm{$0.9201^*$}}    & \multicolumn{1}{l|}{\bm{$0.002577^*$}}    & \multicolumn{1}{l}{\bm{$0.9519^*$}}       & \multicolumn{1}{l|}{\bm{$0.020913^*$}}    & \multicolumn{1}{c}{\bm{$0.9331^*$}}           & \bm{$0.002220^*$} \\  \bottomrule 
    \end{tabular}
    \end{table*}

\section{Experiment Results}
\subsection{Performance Comparison}
Table~\ref{tab:main_result} shows the results of all methods. DTRN achieves the best metrics on all tasks in both the E-commerce and Tenrec datasets, which verifies the effectiveness of DTRN. There are some insightful findings from the results. 
(1) The proposed DTRN beats all baselines on all tasks. Compared with methods of behavior sequence modeling, DTRN achieves more fine-grained interest representation for each task and behavior sequence pair through the Transformer network and hypernetwork. When combined with existing MTL methods in RS, DTRN can boost performance significantly. Both the proposed TIM and TRM module work and ease the learning of the following MTL model. This demonstrates that task-specific bottom representation can alleviate the negative transfer to some extent. 
(2) The performance can be significantly improved by using methods that model multiple types of behavior sequences. For example, methods of modeling multiple types of behavior sequences ATRANK, DFN, and DMT obtain improvement over the methods of modeling the single type of behavior sequence YouTubenet, DIN, DIEN. It reveals that the single type of behavior sequence is not sufficient for describing the user's interest. Exploiting multiple types of behavior sequences can comprehensively capture the user's various interests from a different perspective. 
(3) The negative transfer effects on the shared bottom representation can limit the effectiveness of gating-based parameter-sharing mechanisms. This is evident in the performance of MTL methods in RS, such as MMoE, PLE, AITM, and M2M, which exhibit similar and relatively poor results. SmallHeads which focuses on improving the generalization of the task-shared bottom representation by introducing extra self-auxiliary losses gains improvements over those MTL methods. The result shows the improvement on the bottom representation is reasonable and can boost the MTL performance.
(4) Among methods that take multiple types of behavior sequences, the interest representation extracted by DMT gains the best performance. We think the reason is that DMT allocates individual Transformer encoder-decoder for each behavior sequence respectively, which achieves behavior sequence-specific interest modeling. On the contrary, ATRank directly learns the relatedness of the stitched multiple types of behavior sequences and has the lowest performance. Such results demonstrate the necessity of fine-grained interest representation.

\subsection{Ablation Studies}
We conduct ablation studies on the E-commerce dataset to analyze the effectiveness of DTRN's components.

\subsubsection{Effectiveness of Components in DTRN}

Firstly, we investigate how the TIM and TRM module influence the performance of DTRN and demonstrate the results in Table~\ref{tab:ablation_component}. The ablation experiments are conducted based on the baseline DMT. We integrate TIM with DMT by concatenating the extracting task-specific interest with sparse ID embeddings as partial task-specific bottom representation for each task. When combining TRM with DMT, we set different task-specific representation refinement networks for each task and make the task-shared interest together with sparse ID embeddings serve as input. 

From Table~\ref{tab:ablation_component}, we can find that integrating DMT either with TIM or TRM can outperform the DMT on all tasks, which indicates that full or even partial task-specific bottom representation benefits the MTL. The designed TIM module can capture the fine-grained correlation difference between different tasks and different behavior sequences, reflected as task-specific interest, and gains performance promotion. For the TRM, it achieves the task-related context-aware representation refinement by applying separate refinement network for each task and also increases the metrics. The result of DTRN shows that task-specific interest is an important information of context and the vital refinement operation should act on the task-specific raw bottom representation.

\begin{table}[h]
    \renewcommand\arraystretch{0.85}
    \caption{AUC of TRM and TIM.}\label{tab:ablation_component}
    \vspace{-0.3cm}
    \begin{tabular}{l|c|c|c|c}
    \toprule 
    \multicolumn{1}{c|}{Method}        & \multicolumn{1}{c}{watch}            & \multicolumn{1}{c}{click}            & \multicolumn{1}{c}{entering}         & \multicolumn{1}{c}{conversion}       \\ \midrule 
    \multicolumn{1}{c|}{DMT}                            & \multicolumn{1}{c}{0.7337}            & \multicolumn{1}{c}{0.7245}            & \multicolumn{1}{c}{0.8240}            & \multicolumn{1}{c}{0.8691}            \\  
    \multicolumn{1}{c|}{+TIM}                           & \multicolumn{1}{c}{0.7339}            & \multicolumn{1}{c}{0.7255}            & \multicolumn{1}{c}{0.8264}            & \multicolumn{1}{c}{0.8786}            \\ 
    \multicolumn{1}{c|}{+TRM}                           & \multicolumn{1}{c}{0.7339}            & \multicolumn{1}{c}{0.7257}            & \multicolumn{1}{c}{0.8266}            & \multicolumn{1}{c}{0.8724}            \\
    \multicolumn{1}{c|}{DTRN}                           & \multicolumn{1}{c}{\textbf{0.7346}}   & \multicolumn{1}{c}{\textbf{0.7271}}   & \multicolumn{1}{c}{\textbf{0.8283}}   & \multicolumn{1}{c}{\textbf{0.8828}}   \\ \bottomrule 
    \end{tabular}
\end{table}

\subsubsection{The Way of Injecting Conditional Information into Transformer}

In this section, we do experiments to explore where should the conditional information generated by hypernetwork be injected into. According to the architecture of Transformer, we divide it into QKV, FFN-1, FFN-2, and LN four components. The QKV component generates the "query", "key", and "value" through three separate matrices $W^Q, W^K, W^V$. We inject the conditional information into QKV through the residual mechanism. Specifically, $W_{t,b}^Q=W^Q\odot(1.0+W_{t,b})$, where $W_{t,b}$ is reshaped from the generated parameters by the hypernetwork. $W_{t,b}^Q$ is the actual matrix for generating "query". $W^K, W^V$ hold the same procedure with respective hypernetwork. FFN-1, and FFN-2 represent the first and second layer of the Position-wise Feed-Forward Network. The conditional information will be injected into the FFN's parameters $W_1, b_1$ and $W_2, b_2$. The injection way is the same as QKV through the residual mechanism. LN indicates the layer normalization in Transformer. LN injected with conditional information is our proposed CLN. 

The experiment results are shown in Table~\ref{tab:ablation_tim}. We can see that all the ways have the ability to extract task-specific interest, alleviate compromised task-shared bottom representation, and gain improvement over DMT. Concretely, the best result of LN (DTRN) compared with variants QKV, FFN-1, and FFN-2 highlight the critical role of layer normalization in Transformer, which is consistent with the conclusion in NLP~\cite{mahabadi2021parameter}. Intuitively and mathematically, layer normalization is easier to achieve the representation specificity by scaling, shifting, and ignoring the middle hidden representation.

\begin{table}[h]
    \renewcommand\arraystretch{0.85}
    \caption{AUC of implementing Conditional Transformer.}
    \label{tab:ablation_tim}
    \vspace{-0.2cm}
    \begin{tabular}{cl|c|c|c|c}
    \toprule
    \multicolumn{2}{c|}{Method}        & \multicolumn{1}{c}{watch}              & \multicolumn{1}{c}{click}             & \multicolumn{1}{c}{entering}             & \multicolumn{1}{c}{conversion} \\ \midrule 
    \multicolumn{2}{c|}{DMT}           & \multicolumn{1}{c}{0.7337}             & \multicolumn{1}{c}{0.7245}            & \multicolumn{1}{c}{0.8240}               & \multicolumn{1}{c}{0.8691}  \\
    \multicolumn{2}{c|}{QKV}           & \multicolumn{1}{c}{0.7348}             & \multicolumn{1}{c}{0.7251}            & \multicolumn{1}{c}{0.8265}               & \multicolumn{1}{c}{0.8786}  \\ 
    \multicolumn{2}{c|}{FFN-1}         & \multicolumn{1}{c}{\textbf{0.7352}}    & \multicolumn{1}{c}{0.7266}            & \multicolumn{1}{c}{0.8280}               & \multicolumn{1}{c}{0.8707}  \\ 
    \multicolumn{2}{c|}{FFN-2}         & \multicolumn{1}{c}{0.7348}             & \multicolumn{1}{c}{0.7251}            & \multicolumn{1}{c}{0.8271}               & \multicolumn{1}{c}{0.8715}  \\
    \multicolumn{2}{c|}{LN (DTRN)}     & \multicolumn{1}{c}{0.7346}             & \multicolumn{1}{c}{\textbf{0.7271}}   & \multicolumn{1}{c}{\textbf{0.8283}}      & \multicolumn{1}{c}{\textbf{0.8828}} \\  \bottomrule 
    \end{tabular}
\end{table}

\subsubsection{Combining DTRN with Different MTL Models}
Since we focus on the task-specific bottom representation which is orthogonal to existing multi-task learning models. We show its effectiveness by combining DTRN with different MTL models in RS. The result in Table~\ref{tab:ablation_dtrn_mtl} shows the task-specific bottom representation output by DTRN gives each task stronger ability to acquire its desired representation, alleviating the conflict among tasks, and boosting the performance of all MTL models. What's more, AITM outperforms ShareBottom, MMoE, and PLE when combined with DTRN. This indicates that appropriate network architecture is also an important factor in MTL and can reinforce its own advantage with task-specific bottom representation.

\begin{table}[h]
    \renewcommand\arraystretch{0.85}
    \caption{AUC of combining DTRN with different MTL models.}\label{tab:ablation_dtrn_mtl}
    \vspace{-0.3cm}
    \begin{tabular}{cl|c|c|c|c}
    \toprule
    \multicolumn{2}{c|}{Method}             & \multicolumn{1}{c}{watch}              & \multicolumn{1}{c}{click}             & \multicolumn{1}{c}{entering}             & \multicolumn{1}{c}{conversion} \\  \midrule 
    \multicolumn{2}{c|}{ShareBottom}        & \multicolumn{1}{c}{0.7338}             & \multicolumn{1}{c}{0.7242}            & \multicolumn{1}{c}{0.8235}               & \multicolumn{1}{c}{0.8662}  \\
    \multicolumn{2}{c|}{ShareBottom+DTRN}   & \multicolumn{1}{c}{\textbf{0.7348}}    & \multicolumn{1}{c}{\textbf{0.7270}}   & \multicolumn{1}{c}{\textbf{0.8274}}      & \multicolumn{1}{c}{\textbf{0.8713}}  \\ \midrule 
    \multicolumn{2}{c|}{MMoE}               & \multicolumn{1}{c}{0.7338}             & \multicolumn{1}{c}{0.7246}            & \multicolumn{1}{c}{0.8237}               & \multicolumn{1}{c}{0.8726}  \\
    \multicolumn{2}{c|}{MMoE+DTRN}          & \multicolumn{1}{c}{\textbf{0.7348}}    & \multicolumn{1}{c}{\textbf{0.7267}}   & \multicolumn{1}{c}{\textbf{0.8270}}      & \multicolumn{1}{c}{\textbf{0.8751}}  \\ \midrule 
    \multicolumn{2}{c|}{PLE}                & \multicolumn{1}{c}{0.7340}             & \multicolumn{1}{c}{0.7237}            & \multicolumn{1}{c}{0.8231}               & \multicolumn{1}{c}{0.8692}  \\
    \multicolumn{2}{c|}{PLE+DTRN}           & \multicolumn{1}{c}{\textbf{0.7348}}    & \multicolumn{1}{c}{\textbf{0.7268}}   & \multicolumn{1}{c}{\textbf{0.8273}}      & \multicolumn{1}{c}{\textbf{0.8743}}  \\ \midrule 
    \multicolumn{2}{c|}{AITM}               & \multicolumn{1}{c}{0.7337}             & \multicolumn{1}{c}{0.7245}            & \multicolumn{1}{c}{0.8240}               & \multicolumn{1}{c}{0.8691}  \\
    \multicolumn{2}{c|}{AITM+DTRN}          & \multicolumn{1}{c}{\textbf{0.7346}}    & \multicolumn{1}{c}{\textbf{0.7271}}   & \multicolumn{1}{c}{\textbf{0.8283}}      & \multicolumn{1}{c}{\textbf{0.8828}}  \\ \bottomrule 
    \end{tabular}
\end{table}

\subsubsection{Effect of Different Tasks}
In this subsection, we perform ablation study to observe the influence of different tasks. We still use the DMT for comparison. We study the influence of each task by removing each task in turn and present the result in Table~\ref{tab:ablation_task}. We have the following observations. (1) DTRN outperforms DMT on all tasks in all ablation experiments, which indicates DTRN is robust to different task combinations. (2) From the first two row of Table~\ref{tab:ablation_task}, we find the \emph{watch} task benefits the other task greatly. As the performance of \emph{click}, \emph{entering}, \emph{conversion} drops severely when removing \emph{watch} task. The reason is that the \emph{watch} action is the foundation of the other three tasks because you proceed to the next action only when you have watched the micro-video for more than three seconds and know what the displayed micro-video is. Thus, the \emph{watch} task which contains more positive samples provides more meaningful supervisory signals for the other three tasks. (3) The promotion between \emph{watch} and \emph{click} task is unidirectional. Removing \emph{click} task does not affect the performance of \emph{watch} task. (4) The \emph{entering} task is important for the \emph{conversion} task. The reason is that the \emph{conversion} action happens only after the \emph{entering} task. Directly removing the causal relationship between two tasks leads to performance degradation. (5) The influence among tasks is asymmetrical. The gradients from \emph{entering} task benefit the \emph{conversion} task, but the gradients from \emph{conversion} task has no or negative effect on the \emph{entering} task, the same as \emph{watch} and \emph{click} task. (6) Above all, result of Table~\ref{tab:ablation_task} shows the complex relationship among tasks. Task-specific bottom representation is necessary and effective for handling the complexity.

\begin{table}[h]
    \renewcommand\arraystretch{0.85}
    \caption{AUC of removing each task in turn.}\label{tab:ablation_task}
    \vspace{-0.2cm}
    \begin{tabular}{cl|c|c|c|c}
    \toprule
    \multicolumn{2}{c|}{Method}                     & \multicolumn{1}{c}{watch}          & \multicolumn{1}{c}{click}         & \multicolumn{1}{c}{entering}         & \multicolumn{1}{c}{conversion} \\ \midrule 
    \multirow{2}{*}{-watch}         & DMT           & \multicolumn{1}{c}{-}             & \multicolumn{1}{c}{0.7153}        & \multicolumn{1}{c}{0.8196}           & \multicolumn{1}{c}{0.8631}  \\
                                    & DTRN          & \multicolumn{1}{c}{-}             & \multicolumn{1}{c}{0.7170}        & \multicolumn{1}{c}{0.8221}           & \multicolumn{1}{c}{0.8780}  \\ \midrule 
    \multirow{2}{*}{-click}         & DMT           & \multicolumn{1}{c}{0.7331}         & \multicolumn{1}{c}{-}            & \multicolumn{1}{c}{0.8236}           & \multicolumn{1}{c}{0.8649}  \\
                                    & DTRN          & \multicolumn{1}{c}{0.7349}         & \multicolumn{1}{c}{-}            & \multicolumn{1}{c}{0.8282}           & \multicolumn{1}{c}{0.8811}  \\ \midrule 
    \multirow{2}{*}{-entering}      & DMT           & \multicolumn{1}{c}{0.7327}         & \multicolumn{1}{c}{0.7239}        & \multicolumn{1}{c}{-}               & \multicolumn{1}{c}{0.8652}  \\
                                    & DTRN          & \multicolumn{1}{c}{0.7343}         & \multicolumn{1}{c}{0.7252}        & \multicolumn{1}{c}{-}               & \multicolumn{1}{c}{0.8703}  \\ \midrule 
    \multirow{2}{*}{-conversion}    & DMT           & \multicolumn{1}{c}{0.7327}         & \multicolumn{1}{c}{0.7246}        & \multicolumn{1}{c}{0.8242}           & \multicolumn{1}{c}{-}  \\
                                    & DTRN          & \multicolumn{1}{c}{0.7343}         & \multicolumn{1}{c}{0.7270}        & \multicolumn{1}{c}{0.8284}           & \multicolumn{1}{c}{-}  \\ \bottomrule 
    \end{tabular}
\end{table}

\begin{figure}[h]
    \centering
    \includegraphics[width=0.9\linewidth]{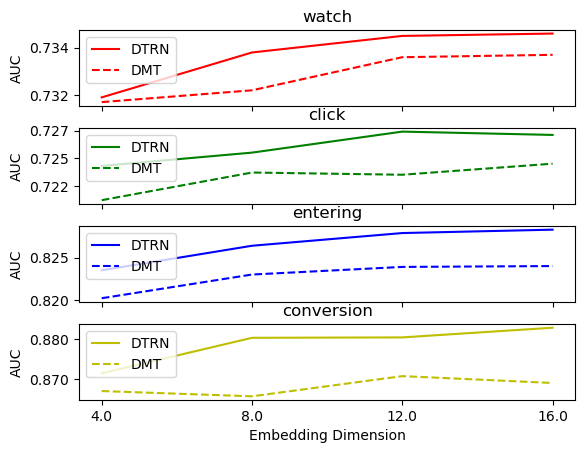}
     \vspace{-0.2cm}
    \caption{AUC of different embedding dimensions.}
    \label{fig:ablation_emb_dim}
\end{figure}

\subsubsection{Hyperparamter: Embedding Dimension}
For hyperparameter, we study the effect of embedding dimension. As Figure~\ref{fig:ablation_emb_dim} shows, DTRN always gains promotion over DMT regardless of the embedding dimension. Another observation is that the sparser the task (i.e. less positive samples), the more promotion. We think the reason is that the magnitude of the sparser task's gradient is smaller and will be influenced by the other tasks heavily. We think that such influence can enhance its performance but the task-shared bottom representation gives a limited promotion upper bound. As above analysis, influence among tasks is not always positive, task-specific bottom representation makes each task have the chance to customize its own representation and hence boosts the performance. Especially, customized representation benefits the sparser tasks that are difficult to train notoriously more.

\subsection{Case Study}
\subsubsection{Task-specific Interest}
\begin{figure}[ht]
    \centering
    \includegraphics[width=0.8\linewidth]{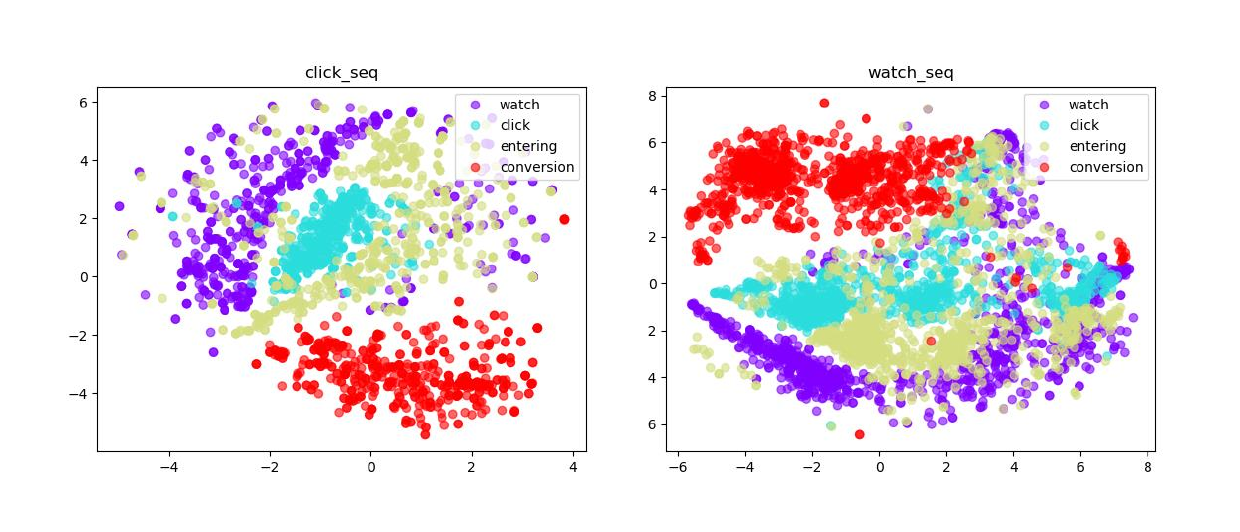}
    \vspace{-0.2cm}
    \caption{Visualization of Task-Specific Interest using t-SNE.}
    \label{fig:seq_task_interest}
\end{figure}

For understanding the TIM module, we visualize the task-specific interest $interest_i$ in Eq.~\eqref{ts_interest}. We select the watch\_seq and click\_seq for visualization. For both behavior sequences, we first randomly select 1000 instances that satisfy the length of the corresponding behavior sequence is more than 40. Then, we feed the 1000 instances into the DTRN to get $4\times1000$ task-specific interest vectors for each behavior sequence respectively. Finally, we visualize these task-specific interest vectors with t-SNE~\cite{van2008visualizing} in Figure~\ref{fig:seq_task_interest}, and we can find that task-specific interest is clearly distinguished. The result shows that the TIM module based on hypernetwork can capture correlation difference between tasks and behavior sequences, and then finish the extraction of the fine-grained interest. 

\begin{figure}[h]
    \centering
    \includegraphics[width=0.6\linewidth]{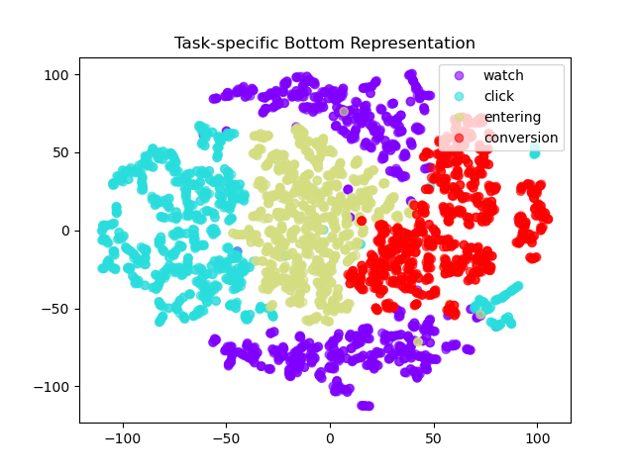}
    \vspace{-0.2cm}
    \caption{Visualization the task-specific representation.}
    \label{fig:task_sepcific_representation}
\end{figure}

\subsubsection{Task-specific Bottom Representation}
Finally, we visualize the task-specific bottom representation, which is the output of the TRM module (i.e. $r_i$ in Eq.~\eqref{trm}.). We randomly select 1000 instances from the test dataset and feed them into the DTRN to get the task-specific bottom representation. The result is shown in Figure~\ref{fig:task_sepcific_representation}, we can see that the representation of each task can be completely separated. This result shows that both our designed DTRN can introduce the task prior knowledge into the bottom representation and obtain the task-specific bottom representation for each task.

\subsection{Online Results}
We conduct A/B test in the online display advertising system to measure the benefits of DTRN compared with the online baseline DMT(MMoE). Both methods are allocated with 10\% serving traffic for 30 days. Table~\ref{tab:online_result} shows the relative promotion of four corresponding objectives. This is a significant improvement in an online display advertising system and proves the effectiveness of DTRN. 

\begin{table}[h]
    \caption{A/B Test of DTRN compared to DMT.}\label{tab:online_result}
    \vspace{-0.4cm}
    \setlength{\tabcolsep}{8mm}
    \begin{tabular}{c|c}
    \toprule
    Indicators            & Accumulated Gains   \\ \midrule 
    Watch Rate            & +0.63\% \\ 
    Click Rate            & +1.00\%  \\ 
    Entering Rate         & +2.23\%  \\
    Revenue Per Mille     & +1.00\%  \\ \bottomrule  
    \end{tabular}
\end{table}

\section{Conclusion}
In this paper, we propose the Deep Task-Specific Bottom Representation Network for the MTL in RS. DTRN which contains TIM and TRM modules aims at generating task-specific bottom representation. 
We conduct offline/online experiments to verify the effectiveness of the DTRN. Finally, we do some ablation studies and visualization to show the correctness of DTRN's component. 

\section{ACKNOWLEDGMENTS}
The work was supported by Alibaba Group through Alibaba Innovative Research Program. Defu Lian was supported by grants from the National Natural Science Foundation of China (No. 62022077 and 61976198).

\clearpage

\bibliographystyle{ACM-Reference-Format}
\balance
\bibliography{sample-base}


\begin{thebibliography}{45}


\ifx \showCODEN    \undefined \def \showCODEN     #1{\unskip}     \fi
\ifx \showDOI      \undefined \def \showDOI       #1{#1}\fi
\ifx \showISBNx    \undefined \def \showISBNx     #1{\unskip}     \fi
\ifx \showISBNxiii \undefined \def \showISBNxiii  #1{\unskip}     \fi
\ifx \showISSN     \undefined \def \showISSN      #1{\unskip}     \fi
\ifx \showLCCN     \undefined \def \showLCCN      #1{\unskip}     \fi
\ifx \shownote     \undefined \def \shownote      #1{#1}          \fi
\ifx \showarticletitle \undefined \def \showarticletitle #1{#1}   \fi
\ifx \showURL      \undefined \def \showURL       {\relax}        \fi
\providecommand\bibfield[2]{#2}
\providecommand\bibinfo[2]{#2}
\providecommand\natexlab[1]{#1}
\providecommand\showeprint[2][]{arXiv:#2}

\bibitem[Caruana(1997)]%
        {caruana1997multitask}
\bibfield{author}{\bibinfo{person}{Rich Caruana}.}
  \bibinfo{year}{1997}\natexlab{}.
\newblock \showarticletitle{Multitask learning}.
\newblock \bibinfo{journal}{\emph{Machine learning}} \bibinfo{volume}{28},
  \bibinfo{number}{1} (\bibinfo{year}{1997}), \bibinfo{pages}{41--75}.
\newblock


\bibitem[Chen et~al\mbox{.}(2022)]%
        {chen2022fast}
\bibfield{author}{\bibinfo{person}{Jin Chen}, \bibinfo{person}{Defu Lian},
  \bibinfo{person}{Binbin Jin}, \bibinfo{person}{Xu Huang},
  \bibinfo{person}{Kai Zheng}, {and} \bibinfo{person}{Enhong Chen}.}
  \bibinfo{year}{2022}\natexlab{}.
\newblock \showarticletitle{Fast variational autoencoder with inverted
  multi-index for collaborative filtering}. In
  \bibinfo{booktitle}{\emph{Proceedings of the ACM Web Conference 2022}}.
  \bibinfo{pages}{1944--1954}.
\newblock


\bibitem[Chen et~al\mbox{.}(2019)]%
        {chen2019behavior}
\bibfield{author}{\bibinfo{person}{Qiwei Chen}, \bibinfo{person}{Huan Zhao},
  \bibinfo{person}{Wei Li}, \bibinfo{person}{Pipei Huang}, {and}
  \bibinfo{person}{Wenwu Ou}.} \bibinfo{year}{2019}\natexlab{}.
\newblock \showarticletitle{Behavior sequence transformer for e-commerce
  recommendation in alibaba}. In \bibinfo{booktitle}{\emph{Proceedings of the
  1st International Workshop on Deep Learning Practice for High-Dimensional
  Sparse Data}}. \bibinfo{pages}{1--4}.
\newblock


\bibitem[Chen et~al\mbox{.}(2021)]%
        {chen2021boosting}
\bibfield{author}{\bibinfo{person}{Xiaokai Chen}, \bibinfo{person}{Xiaoguang
  Gu}, {and} \bibinfo{person}{Libo Fu}.} \bibinfo{year}{2021}\natexlab{}.
\newblock \showarticletitle{Boosting share routing for multi-task learning}. In
  \bibinfo{booktitle}{\emph{Companion Proceedings of the Web Conference 2021}}.
  \bibinfo{pages}{372--379}.
\newblock


\bibitem[Chen et~al\mbox{.}(2018)]%
        {chen2018gradnorm}
\bibfield{author}{\bibinfo{person}{Zhao Chen}, \bibinfo{person}{Vijay
  Badrinarayanan}, \bibinfo{person}{Chen-Yu Lee}, {and} \bibinfo{person}{Andrew
  Rabinovich}.} \bibinfo{year}{2018}\natexlab{}.
\newblock \showarticletitle{Gradnorm: Gradient normalization for adaptive loss
  balancing in deep multitask networks}. In
  \bibinfo{booktitle}{\emph{International conference on machine learning}}.
  PMLR, \bibinfo{pages}{794--803}.
\newblock


\bibitem[Covington et~al\mbox{.}(2016)]%
        {covington2016deep}
\bibfield{author}{\bibinfo{person}{Paul Covington}, \bibinfo{person}{Jay
  Adams}, {and} \bibinfo{person}{Emre Sargin}.}
  \bibinfo{year}{2016}\natexlab{}.
\newblock \showarticletitle{Deep neural networks for youtube recommendations}.
  In \bibinfo{booktitle}{\emph{Proceedings of the 10th ACM conference on
  recommender systems}}. \bibinfo{pages}{191--198}.
\newblock


\bibitem[Ding et~al\mbox{.}(2021)]%
        {ding2021mssm}
\bibfield{author}{\bibinfo{person}{Ke Ding}, \bibinfo{person}{Xin Dong},
  \bibinfo{person}{Yong He}, \bibinfo{person}{Lei Cheng},
  \bibinfo{person}{Chilin Fu}, \bibinfo{person}{Zhaoxin Huan},
  \bibinfo{person}{Hai Li}, \bibinfo{person}{Tan Yan}, \bibinfo{person}{Liang
  Zhang}, \bibinfo{person}{Xiaolu Zhang}, {et~al\mbox{.}}}
  \bibinfo{year}{2021}\natexlab{}.
\newblock \showarticletitle{MSSM: a multiple-level sparse sharing model for
  efficient multi-task learning}. In \bibinfo{booktitle}{\emph{Proceedings of
  the 44th International ACM SIGIR Conference on Research and Development in
  Information Retrieval}}. \bibinfo{pages}{2237--2241}.
\newblock


\bibitem[Gao et~al\mbox{.}(2019)]%
        {gao2019learning}
\bibfield{author}{\bibinfo{person}{Chen Gao}, \bibinfo{person}{Xiangnan He},
  \bibinfo{person}{Dahua Gan}, \bibinfo{person}{Xiangning Chen},
  \bibinfo{person}{Fuli Feng}, \bibinfo{person}{Yong Li},
  \bibinfo{person}{Tat-Seng Chua}, \bibinfo{person}{Lina Yao},
  \bibinfo{person}{Yang Song}, {and} \bibinfo{person}{Depeng Jin}.}
  \bibinfo{year}{2019}\natexlab{}.
\newblock \showarticletitle{Learning to recommend with multiple cascading
  behaviors}.
\newblock \bibinfo{journal}{\emph{IEEE transactions on knowledge and data
  engineering}} \bibinfo{volume}{33}, \bibinfo{number}{6}
  (\bibinfo{year}{2019}), \bibinfo{pages}{2588--2601}.
\newblock


\bibitem[Grbovic and Cheng(2018)]%
        {grbovic2018real}
\bibfield{author}{\bibinfo{person}{Mihajlo Grbovic} {and}
  \bibinfo{person}{Haibin Cheng}.} \bibinfo{year}{2018}\natexlab{}.
\newblock \showarticletitle{Real-time personalization using embeddings for
  search ranking at airbnb}. In \bibinfo{booktitle}{\emph{Proceedings of the
  24th ACM SIGKDD international conference on knowledge discovery \& data
  mining}}. \bibinfo{pages}{311--320}.
\newblock


\bibitem[Gu et~al\mbox{.}(2020)]%
        {gu2020deep}
\bibfield{author}{\bibinfo{person}{Yulong Gu}, \bibinfo{person}{Zhuoye Ding},
  \bibinfo{person}{Shuaiqiang Wang}, \bibinfo{person}{Lixin Zou},
  \bibinfo{person}{Yiding Liu}, {and} \bibinfo{person}{Dawei Yin}.}
  \bibinfo{year}{2020}\natexlab{}.
\newblock \showarticletitle{Deep multifaceted transformers for multi-objective
  ranking in large-scale e-commerce recommender systems}. In
  \bibinfo{booktitle}{\emph{Proceedings of the 29th ACM International
  Conference on Information \& Knowledge Management}}.
  \bibinfo{pages}{2493--2500}.
\newblock


\bibitem[Ha et~al\mbox{.}(2016)]%
        {ha2016hypernetworks}
\bibfield{author}{\bibinfo{person}{David Ha}, \bibinfo{person}{Andrew Dai},
  {and} \bibinfo{person}{Quoc~V Le}.} \bibinfo{year}{2016}\natexlab{}.
\newblock \showarticletitle{Hypernetworks}.
\newblock \bibinfo{journal}{\emph{arXiv preprint arXiv:1609.09106}}
  (\bibinfo{year}{2016}).
\newblock


\bibitem[Huang et~al\mbox{.}(2019)]%
        {huang2019fibinet}
\bibfield{author}{\bibinfo{person}{Tongwen Huang}, \bibinfo{person}{Zhiqi
  Zhang}, {and} \bibinfo{person}{Junlin Zhang}.}
  \bibinfo{year}{2019}\natexlab{}.
\newblock \showarticletitle{FiBiNET: combining feature importance and bilinear
  feature interaction for click-through rate prediction}. In
  \bibinfo{booktitle}{\emph{Proceedings of the 13th ACM Conference on
  Recommender Systems}}. \bibinfo{pages}{169--177}.
\newblock


\bibitem[Kendall et~al\mbox{.}(2018)]%
        {kendall2018multi}
\bibfield{author}{\bibinfo{person}{Alex Kendall}, \bibinfo{person}{Yarin Gal},
  {and} \bibinfo{person}{Roberto Cipolla}.} \bibinfo{year}{2018}\natexlab{}.
\newblock \showarticletitle{Multi-task learning using uncertainty to weigh
  losses for scene geometry and semantics}. In
  \bibinfo{booktitle}{\emph{Proceedings of the IEEE conference on computer
  vision and pattern recognition}}. \bibinfo{pages}{7482--7491}.
\newblock


\bibitem[Lian et~al\mbox{.}(2020a)]%
        {lian2020personalized}
\bibfield{author}{\bibinfo{person}{Defu Lian}, \bibinfo{person}{Qi Liu}, {and}
  \bibinfo{person}{Enhong Chen}.} \bibinfo{year}{2020}\natexlab{a}.
\newblock \showarticletitle{Personalized ranking with importance sampling}. In
  \bibinfo{booktitle}{\emph{Proceedings of The Web Conference 2020}}.
  \bibinfo{pages}{1093--1103}.
\newblock


\bibitem[Lian et~al\mbox{.}(2020b)]%
        {lian2020lightrec}
\bibfield{author}{\bibinfo{person}{Defu Lian}, \bibinfo{person}{Haoyu Wang},
  \bibinfo{person}{Zheng Liu}, \bibinfo{person}{Jianxun Lian},
  \bibinfo{person}{Enhong Chen}, {and} \bibinfo{person}{Xing Xie}.}
  \bibinfo{year}{2020}\natexlab{b}.
\newblock \showarticletitle{Lightrec: A memory and search-efficient recommender
  system}. In \bibinfo{booktitle}{\emph{Proceedings of The Web Conference
  2020}}. \bibinfo{pages}{695--705}.
\newblock


\bibitem[Lin et~al\mbox{.}(2019)]%
        {lin2019pareto}
\bibfield{author}{\bibinfo{person}{Xiao Lin}, \bibinfo{person}{Hongjie Chen},
  \bibinfo{person}{Changhua Pei}, \bibinfo{person}{Fei Sun},
  \bibinfo{person}{Xuanji Xiao}, \bibinfo{person}{Hanxiao Sun},
  \bibinfo{person}{Yongfeng Zhang}, \bibinfo{person}{Wenwu Ou}, {and}
  \bibinfo{person}{Peng Jiang}.} \bibinfo{year}{2019}\natexlab{}.
\newblock \showarticletitle{A pareto-efficient algorithm for multiple objective
  optimization in e-commerce recommendation}. In
  \bibinfo{booktitle}{\emph{Proceedings of the 13th ACM Conference on
  recommender systems}}. \bibinfo{pages}{20--28}.
\newblock


\bibitem[Liu et~al\mbox{.}(2010)]%
        {liu2010unifying}
\bibfield{author}{\bibinfo{person}{Nathan~N Liu}, \bibinfo{person}{Evan~W
  Xiang}, \bibinfo{person}{Min Zhao}, {and} \bibinfo{person}{Qiang Yang}.}
  \bibinfo{year}{2010}\natexlab{}.
\newblock \showarticletitle{Unifying explicit and implicit feedback for
  collaborative filtering}. In \bibinfo{booktitle}{\emph{Proceedings of the
  19th ACM international conference on Information and knowledge management}}.
  \bibinfo{pages}{1445--1448}.
\newblock


\bibitem[Liu et~al\mbox{.}(2019)]%
        {liu2019end}
\bibfield{author}{\bibinfo{person}{Shikun Liu}, \bibinfo{person}{Edward Johns},
  {and} \bibinfo{person}{Andrew~J Davison}.} \bibinfo{year}{2019}\natexlab{}.
\newblock \showarticletitle{End-to-end multi-task learning with attention}. In
  \bibinfo{booktitle}{\emph{Proceedings of the IEEE/CVF conference on computer
  vision and pattern recognition}}. \bibinfo{pages}{1871--1880}.
\newblock


\bibitem[Ma et~al\mbox{.}(2019)]%
        {ma2019snr}
\bibfield{author}{\bibinfo{person}{Jiaqi Ma}, \bibinfo{person}{Zhe Zhao},
  \bibinfo{person}{Jilin Chen}, \bibinfo{person}{Ang Li},
  \bibinfo{person}{Lichan Hong}, {and} \bibinfo{person}{Ed~H Chi}.}
  \bibinfo{year}{2019}\natexlab{}.
\newblock \showarticletitle{Snr: Sub-network routing for flexible parameter
  sharing in multi-task learning}. In \bibinfo{booktitle}{\emph{Proceedings of
  the AAAI Conference on Artificial Intelligence}}, Vol.~\bibinfo{volume}{33}.
  \bibinfo{pages}{216--223}.
\newblock


\bibitem[Ma et~al\mbox{.}(2018b)]%
        {ma2018modeling}
\bibfield{author}{\bibinfo{person}{Jiaqi Ma}, \bibinfo{person}{Zhe Zhao},
  \bibinfo{person}{Xinyang Yi}, \bibinfo{person}{Jilin Chen},
  \bibinfo{person}{Lichan Hong}, {and} \bibinfo{person}{Ed~H Chi}.}
  \bibinfo{year}{2018}\natexlab{b}.
\newblock \showarticletitle{Modeling task relationships in multi-task learning
  with multi-gate mixture-of-experts}. In \bibinfo{booktitle}{\emph{Proceedings
  of the 24th ACM SIGKDD international conference on knowledge discovery \&
  data mining}}. \bibinfo{pages}{1930--1939}.
\newblock


\bibitem[Ma et~al\mbox{.}(2018a)]%
        {ma2018entire}
\bibfield{author}{\bibinfo{person}{Xiao Ma}, \bibinfo{person}{Liqin Zhao},
  \bibinfo{person}{Guan Huang}, \bibinfo{person}{Zhi Wang},
  \bibinfo{person}{Zelin Hu}, \bibinfo{person}{Xiaoqiang Zhu}, {and}
  \bibinfo{person}{Kun Gai}.} \bibinfo{year}{2018}\natexlab{a}.
\newblock \showarticletitle{Entire space multi-task model: An effective
  approach for estimating post-click conversion rate}. In
  \bibinfo{booktitle}{\emph{The 41st International ACM SIGIR Conference on
  Research \& Development in Information Retrieval}}.
  \bibinfo{pages}{1137--1140}.
\newblock


\bibitem[Mahabadi et~al\mbox{.}(2021)]%
        {mahabadi2021parameter}
\bibfield{author}{\bibinfo{person}{Rabeeh~Karimi Mahabadi},
  \bibinfo{person}{Sebastian Ruder}, \bibinfo{person}{Mostafa Dehghani}, {and}
  \bibinfo{person}{James Henderson}.} \bibinfo{year}{2021}\natexlab{}.
\newblock \showarticletitle{Parameter-efficient multi-task fine-tuning for
  transformers via shared hypernetworks}.
\newblock \bibinfo{journal}{\emph{arXiv preprint arXiv:2106.04489}}
  (\bibinfo{year}{2021}).
\newblock


\bibitem[Pan et~al\mbox{.}(2016)]%
        {pan2016mixed}
\bibfield{author}{\bibinfo{person}{Weike Pan}, \bibinfo{person}{Shanchuan Xia},
  \bibinfo{person}{Zhuode Liu}, \bibinfo{person}{Xiaogang Peng}, {and}
  \bibinfo{person}{Zhong Ming}.} \bibinfo{year}{2016}\natexlab{}.
\newblock \showarticletitle{Mixed factorization for collaborative
  recommendation with heterogeneous explicit feedbacks}.
\newblock \bibinfo{journal}{\emph{Information Sciences}}  \bibinfo{volume}{332}
  (\bibinfo{year}{2016}), \bibinfo{pages}{84--93}.
\newblock


\bibitem[Pi et~al\mbox{.}(2020)]%
        {pi2020search}
\bibfield{author}{\bibinfo{person}{Qi Pi}, \bibinfo{person}{Guorui Zhou},
  \bibinfo{person}{Yujing Zhang}, \bibinfo{person}{Zhe Wang},
  \bibinfo{person}{Lejian Ren}, \bibinfo{person}{Ying Fan},
  \bibinfo{person}{Xiaoqiang Zhu}, {and} \bibinfo{person}{Kun Gai}.}
  \bibinfo{year}{2020}\natexlab{}.
\newblock \showarticletitle{Search-based user interest modeling with lifelong
  sequential behavior data for click-through rate prediction}. In
  \bibinfo{booktitle}{\emph{Proceedings of the 29th ACM International
  Conference on Information \& Knowledge Management}}.
  \bibinfo{pages}{2685--2692}.
\newblock


\bibitem[Tang et~al\mbox{.}(2020)]%
        {tang2020progressive}
\bibfield{author}{\bibinfo{person}{Hongyan Tang}, \bibinfo{person}{Junning
  Liu}, \bibinfo{person}{Ming Zhao}, {and} \bibinfo{person}{Xudong Gong}.}
  \bibinfo{year}{2020}\natexlab{}.
\newblock \showarticletitle{Progressive layered extraction (ple): A novel
  multi-task learning (mtl) model for personalized recommendations}. In
  \bibinfo{booktitle}{\emph{Fourteenth ACM Conference on Recommender Systems}}.
  \bibinfo{pages}{269--278}.
\newblock


\bibitem[Tay et~al\mbox{.}(2021)]%
        {tay2021hypergrid}
\bibfield{author}{\bibinfo{person}{Yi Tay}, \bibinfo{person}{Zhe Zhao},
  \bibinfo{person}{Dara Bahri}, \bibinfo{person}{Don Metzler}, {and}
  \bibinfo{person}{Da-Cheng Juan}.} \bibinfo{year}{2021}\natexlab{}.
\newblock \showarticletitle{Hypergrid transformers: Towards a single model for
  multiple tasks}.
\newblock  (\bibinfo{year}{2021}).
\newblock


\bibitem[Van~der Maaten and Hinton(2008)]%
        {van2008visualizing}
\bibfield{author}{\bibinfo{person}{Laurens Van~der Maaten} {and}
  \bibinfo{person}{Geoffrey Hinton}.} \bibinfo{year}{2008}\natexlab{}.
\newblock \showarticletitle{Visualizing data using t-SNE.}
\newblock \bibinfo{journal}{\emph{Journal of machine learning research}}
  \bibinfo{volume}{9}, \bibinfo{number}{11} (\bibinfo{year}{2008}).
\newblock


\bibitem[Vaswani et~al\mbox{.}(2017)]%
        {vaswani2017attention}
\bibfield{author}{\bibinfo{person}{Ashish Vaswani}, \bibinfo{person}{Noam
  Shazeer}, \bibinfo{person}{Niki Parmar}, \bibinfo{person}{Jakob Uszkoreit},
  \bibinfo{person}{Llion Jones}, \bibinfo{person}{Aidan~N Gomez},
  \bibinfo{person}{{\L}ukasz Kaiser}, {and} \bibinfo{person}{Illia
  Polosukhin}.} \bibinfo{year}{2017}\natexlab{}.
\newblock \showarticletitle{Attention is all you need}.
\newblock \bibinfo{journal}{\emph{Advances in neural information processing
  systems}}  \bibinfo{volume}{30} (\bibinfo{year}{2017}).
\newblock


\bibitem[Wang et~al\mbox{.}(2022)]%
        {wang2022can}
\bibfield{author}{\bibinfo{person}{Yuyan Wang}, \bibinfo{person}{Zhe Zhao},
  \bibinfo{person}{Bo Dai}, \bibinfo{person}{Christopher Fifty},
  \bibinfo{person}{Dong Lin}, \bibinfo{person}{Lichan Hong},
  \bibinfo{person}{Li Wei}, {and} \bibinfo{person}{Ed~H Chi}.}
  \bibinfo{year}{2022}\natexlab{}.
\newblock \showarticletitle{Can Small Heads Help? Understanding and Improving
  Multi-Task Generalization}. In \bibinfo{booktitle}{\emph{Proceedings of the
  ACM Web Conference 2022}}. \bibinfo{pages}{3009--3019}.
\newblock


\bibitem[Wang et~al\mbox{.}(2020)]%
        {wang2020gradient}
\bibfield{author}{\bibinfo{person}{Zirui Wang}, \bibinfo{person}{Yulia
  Tsvetkov}, \bibinfo{person}{Orhan Firat}, {and} \bibinfo{person}{Yuan Cao}.}
  \bibinfo{year}{2020}\natexlab{}.
\newblock \showarticletitle{Gradient vaccine: Investigating and improving
  multi-task optimization in massively multilingual models}.
\newblock \bibinfo{journal}{\emph{arXiv preprint arXiv:2010.05874}}
  (\bibinfo{year}{2020}).
\newblock


\bibitem[Wen et~al\mbox{.}(2021)]%
        {wen2021hierarchically}
\bibfield{author}{\bibinfo{person}{Hong Wen}, \bibinfo{person}{Jing Zhang},
  \bibinfo{person}{Fuyu Lv}, \bibinfo{person}{Wentian Bao},
  \bibinfo{person}{Tianyi Wang}, {and} \bibinfo{person}{Zulong Chen}.}
  \bibinfo{year}{2021}\natexlab{}.
\newblock \showarticletitle{Hierarchically modeling micro and macro behaviors
  via multi-task learning for conversion rate prediction}. In
  \bibinfo{booktitle}{\emph{Proceedings of the 44th International ACM SIGIR
  Conference on Research and Development in Information Retrieval}}.
  \bibinfo{pages}{2187--2191}.
\newblock


\bibitem[Wen et~al\mbox{.}(2020)]%
        {wen2020entire}
\bibfield{author}{\bibinfo{person}{Hong Wen}, \bibinfo{person}{Jing Zhang},
  \bibinfo{person}{Yuan Wang}, \bibinfo{person}{Fuyu Lv},
  \bibinfo{person}{Wentian Bao}, \bibinfo{person}{Quan Lin}, {and}
  \bibinfo{person}{Keping Yang}.} \bibinfo{year}{2020}\natexlab{}.
\newblock \showarticletitle{Entire space multi-task modeling via post-click
  behavior decomposition for conversion rate prediction}. In
  \bibinfo{booktitle}{\emph{Proceedings of the 43rd International ACM SIGIR
  conference on research and development in Information Retrieval}}.
  \bibinfo{pages}{2377--2386}.
\newblock


\bibitem[Wu et~al\mbox{.}(2023)]%
        {wu2023influence}
\bibfield{author}{\bibinfo{person}{Chenwang Wu}, \bibinfo{person}{Defu Lian},
  \bibinfo{person}{Yong Ge}, \bibinfo{person}{Zhihao Zhu}, {and}
  \bibinfo{person}{Enhong Chen}.} \bibinfo{year}{2023}\natexlab{}.
\newblock \showarticletitle{Influence-Driven Data Poisoning for Robust
  Recommender Systems}.
\newblock \bibinfo{journal}{\emph{IEEE Transactions on Pattern Analysis and
  Machine Intelligence}} (\bibinfo{year}{2023}).
\newblock


\bibitem[Wu et~al\mbox{.}(2022)]%
        {wu2022multi}
\bibfield{author}{\bibinfo{person}{Xuyang Wu}, \bibinfo{person}{Alessandro
  Magnani}, \bibinfo{person}{Suthee Chaidaroon}, \bibinfo{person}{Ajit
  Puthenputhussery}, \bibinfo{person}{Ciya Liao}, {and} \bibinfo{person}{Yi
  Fang}.} \bibinfo{year}{2022}\natexlab{}.
\newblock \showarticletitle{A Multi-task Learning Framework for Product Ranking
  with BERT}. In \bibinfo{booktitle}{\emph{Proceedings of the ACM Web
  Conference 2022}}. \bibinfo{pages}{493--501}.
\newblock


\bibitem[Xi et~al\mbox{.}(2021)]%
        {xi2021modeling}
\bibfield{author}{\bibinfo{person}{Dongbo Xi}, \bibinfo{person}{Zhen Chen},
  \bibinfo{person}{Peng Yan}, \bibinfo{person}{Yinger Zhang},
  \bibinfo{person}{Yongchun Zhu}, \bibinfo{person}{Fuzhen Zhuang}, {and}
  \bibinfo{person}{Yu Chen}.} \bibinfo{year}{2021}\natexlab{}.
\newblock \showarticletitle{Modeling the sequential dependence among audience
  multi-step conversions with multi-task learning in targeted display
  advertising}. In \bibinfo{booktitle}{\emph{Proceedings of the 27th ACM SIGKDD
  Conference on Knowledge Discovery \& Data Mining}}.
  \bibinfo{pages}{3745--3755}.
\newblock


\bibitem[Xie et~al\mbox{.}(2021)]%
        {xie2021deep}
\bibfield{author}{\bibinfo{person}{Ruobing Xie}, \bibinfo{person}{Cheng Ling},
  \bibinfo{person}{Yalong Wang}, \bibinfo{person}{Rui Wang},
  \bibinfo{person}{Feng Xia}, {and} \bibinfo{person}{Leyu Lin}.}
  \bibinfo{year}{2021}\natexlab{}.
\newblock \showarticletitle{Deep feedback network for recommendation}. In
  \bibinfo{booktitle}{\emph{Proceedings of the Twenty-Ninth International
  Conference on International Joint Conferences on Artificial Intelligence}}.
  \bibinfo{pages}{2519--2525}.
\newblock


\bibitem[Yu et~al\mbox{.}(2020)]%
        {yu2020gradient}
\bibfield{author}{\bibinfo{person}{Tianhe Yu}, \bibinfo{person}{Saurabh Kumar},
  \bibinfo{person}{Abhishek Gupta}, \bibinfo{person}{Sergey Levine},
  \bibinfo{person}{Karol Hausman}, {and} \bibinfo{person}{Chelsea Finn}.}
  \bibinfo{year}{2020}\natexlab{}.
\newblock \showarticletitle{Gradient surgery for multi-task learning}.
\newblock \bibinfo{journal}{\emph{Advances in Neural Information Processing
  Systems}}  \bibinfo{volume}{33} (\bibinfo{year}{2020}),
  \bibinfo{pages}{5824--5836}.
\newblock


\bibitem[Zhang et~al\mbox{.}(2022b)]%
        {zhang2022leaving}
\bibfield{author}{\bibinfo{person}{Qianqian Zhang}, \bibinfo{person}{Xinru
  Liao}, \bibinfo{person}{Quan Liu}, \bibinfo{person}{Jian Xu}, {and}
  \bibinfo{person}{Bo Zheng}.} \bibinfo{year}{2022}\natexlab{b}.
\newblock \showarticletitle{Leaving No One Behind: A Multi-Scenario Multi-Task
  Meta Learning Approach for Advertiser Modeling}. In
  \bibinfo{booktitle}{\emph{Proceedings of the Fifteenth ACM International
  Conference on Web Search and Data Mining}}. \bibinfo{pages}{1368--1376}.
\newblock


\bibitem[Zhang et~al\mbox{.}(2022a)]%
        {zhang2022picasso}
\bibfield{author}{\bibinfo{person}{Yuanxing Zhang}, \bibinfo{person}{Langshi
  Chen}, \bibinfo{person}{Siran Yang}, \bibinfo{person}{Man Yuan},
  \bibinfo{person}{Huimin Yi}, {et~al\mbox{.}}}
  \bibinfo{year}{2022}\natexlab{a}.
\newblock \showarticletitle{PICASSO: Unleashing the Potential of GPU-centric
  Training for Wide-and-deep Recommender Systems}. In
  \bibinfo{booktitle}{\emph{2022 IEEE 38th International Conference on Data
  Engineering (ICDE)}}. IEEE.
\newblock


\bibitem[Zhang et~al\mbox{.}(2022c)]%
        {zhang2022towards}
\bibfield{author}{\bibinfo{person}{Zhao-Yu Zhang}, \bibinfo{person}{Xiang-Rong
  Sheng}, \bibinfo{person}{Yujing Zhang}, \bibinfo{person}{Biye Jiang},
  \bibinfo{person}{Shuguang Han}, \bibinfo{person}{Hongbo Deng}, {and}
  \bibinfo{person}{Bo Zheng}.} \bibinfo{year}{2022}\natexlab{c}.
\newblock \showarticletitle{Towards Understanding the Overfitting Phenomenon of
  Deep Click-Through Rate Prediction Models}.
\newblock \bibinfo{journal}{\emph{arXiv preprint arXiv:2209.06053}}
  (\bibinfo{year}{2022}).
\newblock


\bibitem[Zhao et~al\mbox{.}(2019)]%
        {zhao2019recommending}
\bibfield{author}{\bibinfo{person}{Zhe Zhao}, \bibinfo{person}{Lichan Hong},
  \bibinfo{person}{Li Wei}, \bibinfo{person}{Jilin Chen},
  \bibinfo{person}{Aniruddh Nath}, \bibinfo{person}{Shawn Andrews},
  \bibinfo{person}{Aditee Kumthekar}, \bibinfo{person}{Maheswaran
  Sathiamoorthy}, \bibinfo{person}{Xinyang Yi}, {and} \bibinfo{person}{Ed
  Chi}.} \bibinfo{year}{2019}\natexlab{}.
\newblock \showarticletitle{Recommending what video to watch next: a multitask
  ranking system}. In \bibinfo{booktitle}{\emph{Proceedings of the 13th ACM
  Conference on Recommender Systems}}. \bibinfo{pages}{43--51}.
\newblock


\bibitem[Zhou et~al\mbox{.}(2018a)]%
        {zhou2018atrank}
\bibfield{author}{\bibinfo{person}{Chang Zhou}, \bibinfo{person}{Jinze Bai},
  \bibinfo{person}{Junshuai Song}, \bibinfo{person}{Xiaofei Liu},
  \bibinfo{person}{Zhengchao Zhao}, \bibinfo{person}{Xiusi Chen}, {and}
  \bibinfo{person}{Jun Gao}.} \bibinfo{year}{2018}\natexlab{a}.
\newblock \showarticletitle{Atrank: An attention-based user behavior modeling
  framework for recommendation}. In \bibinfo{booktitle}{\emph{Thirty-Second
  AAAI Conference on Artificial Intelligence}}.
\newblock


\bibitem[Zhou et~al\mbox{.}(2019)]%
        {zhou2019deep}
\bibfield{author}{\bibinfo{person}{Guorui Zhou}, \bibinfo{person}{Na Mou},
  \bibinfo{person}{Ying Fan}, \bibinfo{person}{Qi Pi}, \bibinfo{person}{Weijie
  Bian}, \bibinfo{person}{Chang Zhou}, \bibinfo{person}{Xiaoqiang Zhu}, {and}
  \bibinfo{person}{Kun Gai}.} \bibinfo{year}{2019}\natexlab{}.
\newblock \showarticletitle{Deep interest evolution network for click-through
  rate prediction}. In \bibinfo{booktitle}{\emph{Proceedings of the AAAI
  conference on artificial intelligence}}, Vol.~\bibinfo{volume}{33}.
  \bibinfo{pages}{5941--5948}.
\newblock


\bibitem[Zhou et~al\mbox{.}(2018b)]%
        {zhou2018deep}
\bibfield{author}{\bibinfo{person}{Guorui Zhou}, \bibinfo{person}{Xiaoqiang
  Zhu}, \bibinfo{person}{Chenru Song}, \bibinfo{person}{Ying Fan},
  \bibinfo{person}{Han Zhu}, \bibinfo{person}{Xiao Ma},
  \bibinfo{person}{Yanghui Yan}, \bibinfo{person}{Junqi Jin},
  \bibinfo{person}{Han Li}, {and} \bibinfo{person}{Kun Gai}.}
  \bibinfo{year}{2018}\natexlab{b}.
\newblock \showarticletitle{Deep interest network for click-through rate
  prediction}. In \bibinfo{booktitle}{\emph{Proceedings of the 24th ACM SIGKDD
  international conference on knowledge discovery \& data mining}}.
  \bibinfo{pages}{1059--1068}.
\newblock


\bibitem[Zou et~al\mbox{.}(2022)]%
        {zou2022automatic}
\bibfield{author}{\bibinfo{person}{Xinyu Zou}, \bibinfo{person}{Zhi Hu},
  \bibinfo{person}{Yiming Zhao}, \bibinfo{person}{Xuchu Ding},
  \bibinfo{person}{Zhongyi Liu}, \bibinfo{person}{Chenliang Li}, {and}
  \bibinfo{person}{Aixin Sun}.} \bibinfo{year}{2022}\natexlab{}.
\newblock \showarticletitle{Automatic Expert Selection for Multi-Scenario and
  Multi-Task Search}.
\newblock \bibinfo{journal}{\emph{arXiv preprint arXiv:2205.14321}}
  (\bibinfo{year}{2022}).
\newblock


\end{thebibliography}
\clearpage
\appendix

\end{document}